\documentclass[journal]{IEEEtran}

\ifCLASSINFOpdf
  \usepackage[pdftex]{graphicx}
  \graphicspath{{./figs/}}
  \DeclareGraphicsExtensions{.pdf,.jpeg,.png}
\else
\fi
\usepackage[cmex10]{amsmath}
\usepackage{mathtools}
\usepackage{amssymb}
\usepackage{algorithm}
\usepackage{algpseudocode}

\usepackage{array}
\PassOptionsToPackage{hyphens}{url}\usepackage{hyperref}

\usepackage{biograph}        
\usepackage{afterpage}
\usepackage{array}
\usepackage{booktabs}

\usepackage{cite}
\usepackage{multiro                                                                                                                                                                                     w} 
\usepackage{rotating}

\usepackage{float}
\usepackage{color}
\usepackage{soul}

\usepackage{colortbl}

\def\por1{\partial}

\newcolumntype{S}{>{\centering\arraybackslash} m{.4\linewidth} }

\makeatletter
  \newcommand\tinyv{\@setfontsize\tinyv{5pt}{7}}
\makeatother

\newlength{\hspacephantom}
\begin{document}
\DeclareGraphicsExtensions{.pdf,.jpeg,.png}

\title{Mean Deviation Similarity Index: Efficient and Reliable Full-Reference Image Quality Evaluator}



\author{Hossein Ziaei Nafchi, Atena Shahkolaei, Rachid Hedjam, \textnormal{and} Mohamed Cheriet, \IEEEmembership{Senior Member,~IEEE} 
\thanks{H. Ziaei Nafchi, A. Shahkolaei and M. Cheriet are with the Synchromedia Laboratory for Multimedia Communication in Telepresence,
\'Ecole de technologie sup\'erieure, Montreal (QC), Canada H3C 1K3; Tel.: +1-514-396-8972; Fax: +1-514-396-8595;
Emails: hossein.zi@synchromedia.ca, atena.shahkolaei.1@ens.etsmtl.ca, mohamed.cheriet@etsmtl.ca}
\thanks{R. Hedjam is with the Department of Geography, McGill University, 805 Sherbrooke Street West, Montreal, QC H3A 2K6, Canada (email: rachid.hedjam@mcgill.ca)}}

\markboth{}%
{}
%


\maketitle

\begin{abstract}
Applications of perceptual image quality assessment (IQA) in image and video processing, such as image acquisition, image compression, image restoration and multimedia communication, have led to the development of many IQA metrics. In this paper, a reliable full reference IQA model is proposed that utilize gradient similarity (GS), chromaticity similarity (CS), and deviation pooling (DP). By considering the shortcomings of the commonly used GS to model human visual system (HVS), a new GS is proposed through a fusion technique that is more likely to follow HVS. We propose an efficient and effective formulation to calculate the joint similarity map of two chromatic channels for the purpose of measuring color changes. In comparison with a commonly used formulation in the literature, the proposed CS map is shown to be more efficient and provide comparable or better quality predictions. Motivated by a recent work that utilizes the standard deviation pooling, a general formulation of the DP is presented in this paper and used to compute a final score from the proposed GS and CS maps. This proposed formulation of DP benefits from the Minkowski pooling and a proposed power pooling as well. The experimental results on six datasets of natural images, a synthetic dataset, and a digitally retouched dataset show that the proposed index provides comparable or better quality predictions than the most recent and competing state-of-the-art IQA metrics in the literature, it is reliable and has low complexity. The MATLAB source code of the proposed metric is available at https://www.mathworks.com/matlabcentral/fileexchange/59809.      
\end{abstract}


%
\IEEEpeerreviewmaketitle


\maketitle

\section{Introduction}
\label{sec:intro}


\IEEEPARstart{A}{utomatic}
image quality assessment (IQA) plays a significant
role in numerous image and video processing applications. IQA is commonly used in image acquisition, image compression, image restoration, multimedia streaming, etc \cite{applications2011}. IQA models (IQAs) mimic the average quality predictions of human observers. Full reference IQAs (FR-IQAs), which fall within the scope of this paper, evaluate the perceptual quality of a distorted image with respect to its reference image. This quality prediction is an easy task for the human visual system (HVS) and the result of the evaluation is reliable. Automatic quality assessment, e.g. objective evaluation, is not an easy task because images may suffer from various types and degrees of distortions. FR-IQAs can be employed to compare two images of the same dynamic range (usually low dynamic range) \cite{SSIM} or different dynamic ranges \cite{TMQI, FSITM}. This paper is dedicated to the IQA for low dynamic range images. 

Among IQAs, the conventional metric mean squared error (MSE) and its variations are widely used because of their simplicity. However, in many situations, MSE does not correlate with the human perception of image fidelity and quality \cite{spm2009}. Because of this limitation, a number of IQAs have been proposed to provide better correlation with the HVS \cite{NQM, SSIM, MSSSIM, IFC, VIF, VSNR, RFSIM, MAD, SVR2010, CPSSIM, IWSSIM, FSIM, GS, SRSIM, GMSD, SFF, VSI, ADD-GSIM, SCQI, Fusion2016}. In general, these better performing metrics measure structural information, luminance information and contrast information in the spatial and frequency domains.   

The most successful IQA models in the literature follow a top-down strategy \cite{Metrics2011}. They calculate a similarity map and use a pooling strategy that converts the values of this similarity map into a single quality score. For example, the luminance, contrast and structural information constitute a similarity map for the popular SSIM index \cite{SSIM}. SSIM then uses average pooling to compute the final similarity score. Different feature maps are used in the literature for calculation of this similarity map. Feature similarity index (FSIM) uses phase congruency and gradient magnitude features. GS \cite{GS} uses a combination of some designated gradient magnitudes and image contrast for this end, while the GMSD \cite{GMSD} uses only the gradient magnitude. SR\_SIM \cite{SRSIM} uses saliency features and gradient magnitude. VSI \cite{VSI} likewise benefits from saliency-based features and gradient magnitude. SVD based features \cite{SVD2006}, features based on the Riesz transform \cite{RFSIM}, features in the wavelet domain \cite{VSNR, wavelet2011, CWSSIM} and sparse features \cite{SFF} are used as well in the literature. Among these features, gradient magnitude is an efficient feature, as shown in \cite{GMSD}. In contrast, phase congruency and visual saliency features in general are not fast enough features to be used. Therefore, the features being used play a significant role in the efficiency of IQAs.   

As we mentioned earlier, the computation of the similarity map is followed by a pooling strategy. The state-of-the-art pooling strategies for perceptual image quality assessment (IQA) are based on the mean and the weighted mean \cite{SSIM, MSSSIM, IWSSIM, GS, FSIM, RFSIM}. They are robust pooling strategies that usually provide a moderate to high performance for different IQAs. Minkowski pooling \cite{spatial2006}, local distortion pooling \cite{spatial2006, pooling2009, MAD}, percentile pooling \cite{percentile2009} and saliency-based pooling \cite{SRSIM, VSI} are other possibilities. Standard deviation (SD) pooling was also proposed and successfully used by GMSD \cite{GMSD}. The image gradients are sensitive to image distortions. Different local structures in a distorted image suffer from different degrees of degradations. This is the motivation that the authors in \cite{GMSD} used to explore the standard variation of the gradient-based local similarity map for overall image quality prediction. In general, features that constitute the similarity map and the pooling strategy are very important factors in designing high performance IQA models.

Here, we propose an IQA model called the mean deviation similarity index (MDSI) that shows very good compromise between prediction accuracy and model complexity. The proposed index is efficient, effective and reliable at the same time. It also shows consistent performance for both natural and synthetic images. The proposed metric follows a top-down strategy. It uses gradient magnitude to measure structural distortions and use chrominance features to measure color distortions. These two similarity maps are then combined to form a gradient-chromaticity similarity map. We then propose a novel deviation pooling strategy and use it to compute the final quality score. Both image gradient \cite{GSSIM, gradient2010, GS, FSIM, GMSD, VSI} and chrominance features \cite{FSIM, VSI} have been already used in the literature. The proposed MDSI uses a new gradient similarity which is more likely to follow HVS. Also, MDSI uses a new chromaticity similarity map which is efficient and shows good performance when used with the proposed metric. The proposed index uses the summation over similarity maps to give independent weights to them. Also, less attention has been paid to the deviation pooling strategy, except for a special case of this type of pooling, namely, standard deviation pooling \cite{GMSD}. We therefore provide a general formulation for the deviation pooling strategy and show its power in the case of the proposed IQA model. In the following, the main contributions of the paper as well as its differences with respect to the previous works are briefly explained. 

Unlike previous researches \cite{GSSIM, gradient2010, GS, FSIM, GMSD, VSI} that use a similar gradient similarity map, a new gradient similarity map is proposed in this paper which is more likely to follow the human visual system (HVS). This statement is supported by visual examples and experimental results.

This paper proposes a new chormaticity similarity map with the following advantages over the previously used chromaticity similarity maps \cite{FSIM, VSI}. Its complexity is lower and it provides slightly better quality predictions when used with the proposed metric.     

Motivated by a previous study that proposed to use standard deviation pooling \cite{GMSD}, we propose a systematic and general formulation of the deviation pooling which has a comprehensive scope.    


The rest of the paper is organized as follows. The proposed mean deviation similarity index is presented in section \ref{MDSI}. Extensive experimental results and discussion on six natural datasets, a synthetic dataset, and a digitally retouched dataset are provided in section \ref{results}. Section \ref{conclusion} presents our conclusions.

\section{Mean Deviation Similarity Index}
\label{MDSI}

The proposed IQA model uses two similarity maps. Image gradient, which is sensitive to structural distortions, is used as the main feature to calculate the first similarity map. Then, color distortions are measured by a chromaticity similarity map. These similarity maps are combined and pooled by a proposed deviation pooling strategy. In this paper, conversion to luminance is done through the following formula: $L = 0.2989R + 0.5870G + 0.1140B$. In addition, two chromaticity channels of a Gaussian color model \cite{invariance2001} are used:     
\begin{equation}
\small
  \ \begin{bmatrix}
H \\ M \end{bmatrix} = \begin{pmatrix}
0.30 & 0.04 & -0.35 \\ 0.34 & -0.6 & 0.17 \end{pmatrix}  \begin{bmatrix} R \\ G \\ B \end{bmatrix}
  \label{operators}
\end{equation}

\subsection{Gradient Similarity}
\label{GSsection}

It is very common that gradient magnitude in the discrete domain is calculated on the basis of some operators that approximate derivatives of the image function using differences. These operators approximate vertical $G_y(\textbf{x})$ and horizontal $G_x(\textbf{x})$ gradients of an image $f(\textbf{x})$ using convolution: $G_x(\textbf{x}) = h_x \ast f(\textbf{x})$ and $G_y(\textbf{x}) = h_y \ast f(\textbf{x})$, where $h_x$ and $h_y$ are horizontal and vertical gradient operators and $\ast$ denotes the convolution. The first derivative magnitude is defined as $G(\textbf{x})=\sqrt{G^2_x(\textbf{x})+G^2_y(\textbf{x})}$. The Sobel operator \cite{sobel}, the Scharr operator, and the Prewitt operator are common gradient operators that approximate first derivatives. Within the proposed IQA model, these operators perform almost the same.      

Through this paper, Prewitt operator is used to compute gradient magnitudes of luminance $L$ channels of reference and distorted images, $\mathcal{R}$ and $\mathcal{D}$. From which, gradient similarity (GS) is computed by the following SSIM induced equation:
\begin{equation}
  \ \text{GS(x)}=\frac{2 G_\mathcal{R}(x) G_\mathcal{D}(x) + C_1} {G_\mathcal{R}^2(x) + G_\mathcal{D}^2(x) + C_1}
  \label{GS}
\end{equation}                    
where, parameter $C_1$ is a constant to control numerical stability. The gradient similarity (GS) is widely used in the literature \cite{GSSIM, gradient2010, GS, FSIM, GMSD, VSI} and its usefulness to measure image distortions was extensively investigated in \cite{GMSD}.

In many scenarios, human visual system (HVS) disagrees with the judgments provided by the GS for structural distortions. In fact, in such a formulation, there is no difference between an added edge to or a removed edge from the distorted image with \textit{respect to the reference image}. An extra edge in $\mathcal{D}$ bring less attention of HVS if its color is close to the relative pixels of that edge in $\mathcal{R}$. Likewise, HVS pays less attention to a removed edge from $\mathcal{R}$ that is replaced with pixels of the same or nearly the same color. In another scenario, suppose that edges are preserved in $\mathcal{D}$ but with different colors than in $\mathcal{R}$. In this case, GS is likely to fail at providing a good judgment ``on the edges". These shortcomings of the GS motivated us to propose a new GS map.

\begin{figure*}[htb]
\scriptsize
\begin{minipage}[b]{0.197\linewidth}
  \centering
  \centerline{\includegraphics[height=2.3cm]{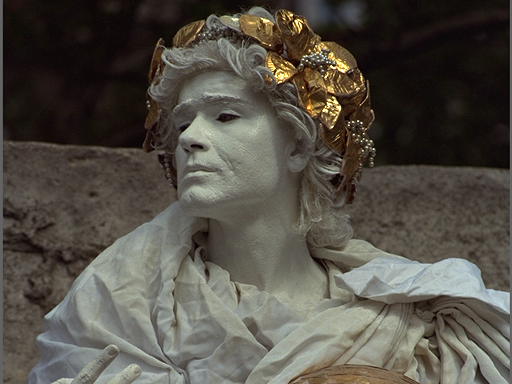}} 
  \vspace{0.15cm}
\centerline{$\mathcal{R}$}
\end{minipage}
\begin{minipage}[b]{0.197\linewidth}
  \centering
  \centerline{\includegraphics[height=2.3cm]{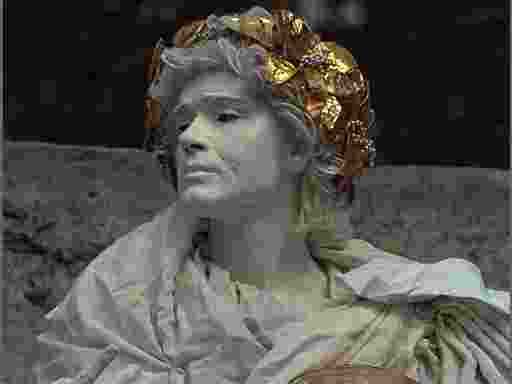}}
  \vspace{0.15cm}
\centerline{$\mathcal{D}$ (JPEG compression)}
\end{minipage}
\begin{minipage}[b]{0.197\linewidth}
  \centering
  \centerline{\fbox{\includegraphics[height=2.3cm]{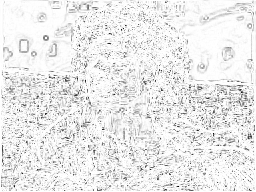}}}
\vspace{0.15cm}
\centerline{GS}
\end{minipage}
\begin{minipage}[b]{0.197\linewidth}
  \centering
  \centerline{\fbox{\includegraphics[height=2.3cm]{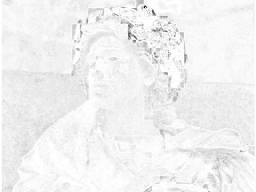}}}
\vspace{0.09cm}
\centerline{$\widehat{\text{CS}}$}
\end{minipage}
\begin{minipage}[b]{.197\linewidth}
  \centering
  \centerline{\fbox{\includegraphics[height=2.3cm]{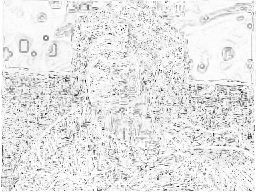}}}
\vspace{0.15cm}
\centerline{GCS ($\alpha = 0.7$)}
\end{minipage}
\\ \\
\begin{minipage}[b]{0.197\linewidth}
  \centering
  \centerline{\includegraphics[height=2.3cm]{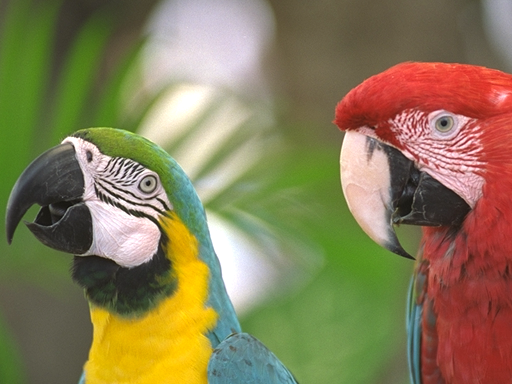}} 
  \vspace{0.15cm}
\centerline{$\mathcal{R}$}
\end{minipage}
\begin{minipage}[b]{0.197\linewidth}
  \centering
  \centerline{\includegraphics[height=2.3cm]{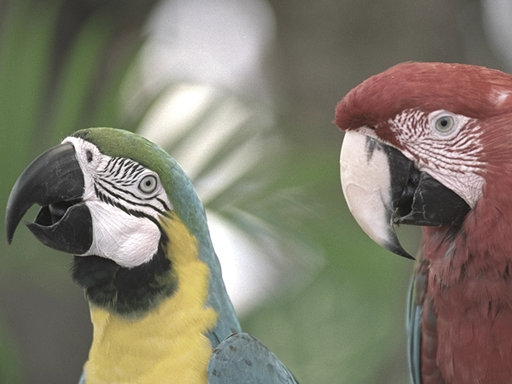}}
  \vspace{0.15cm}
\centerline{$\mathcal{D}$ (color saturation)}
\end{minipage}
\begin{minipage}[b]{0.197\linewidth}
  \centering
  \centerline{\fbox{\includegraphics[height=2.3cm]{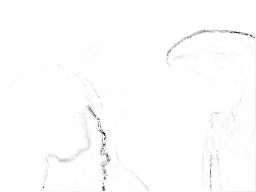}}}
\vspace{0.15cm}
\centerline{GS}
\end{minipage}
\begin{minipage}[b]{0.197\linewidth}
  \centering
  \centerline{\fbox{\includegraphics[height=2.3cm]{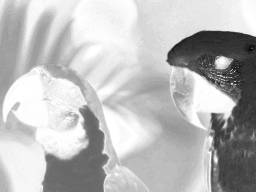}}}
\vspace{0.09cm}
\centerline{$\widehat{\text{CS}}$}
\end{minipage}
\begin{minipage}[b]{.197\linewidth}
  \centering
  \centerline{\fbox{\includegraphics[height=2.3cm]{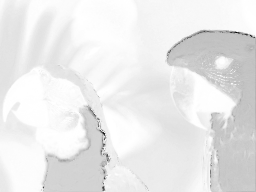}}}
\vspace{0.15cm}
\centerline{GCS ($\alpha = 0.7$)}
\end{minipage}
\caption{Complementary behavior of the gradient similarity (GS) and chromaticity similarity ($\widehat{\text{CS}}$) maps.}
\label{fig:out1}
\end{figure*}

\subsection{The Proposed Gradient Similarity} 
\label{GShat}

The aforementioned shortcomings of the conventional gradient similarity map (equation \ref{GS}) are mainly because $G_\mathcal{R}$ and $G_\mathcal{D}$ are computed independent of each other. In the following, we propose a fusion technique to include the correlation between $\mathcal{R}$ and $\mathcal{D}$ images into computation of the gradient similarity map.   

We fuse the luminance $L$ channels of the $\mathcal{R}$ and $\mathcal{D}$ by a simple averaging: $\mathcal{F}$ = 0.5 $\times$ ($\mathcal{R}$ + $\mathcal{D}$). Two extra GS maps are computed as follows: 
\begin{equation}
  \ \text{GS}_{\mathcal{RF}}\text{(x)}=\frac{2 G_\mathcal{R}(x) G_\mathcal{F}(x) + C_2} {G_\mathcal{R}^2(x) + G_\mathcal{F}^2(x) + C_2}
  \label{GS13}
\end{equation}                    
\begin{equation}
  \ \text{GS}_{\mathcal{DF}}\text{(x)}=\frac{2 G_\mathcal{D}(x) G_\mathcal{F}(x) + C_2} {G_\mathcal{D}^2(x) + G_\mathcal{F}^2(x) + C_2}
  \label{GS23}
\end{equation}                    
where, $G_\mathcal{F}$ is the gradient magnitude of the fused image $\mathcal{F}$, and $C_2$ is used for numerical stability. Note that $G_\mathcal{F} \neq (G_\mathcal{R} + G_\mathcal{D}) / 2$, and that  $\text{GS}_{\mathcal{RF}}\text{(x)}$ and $\text{GS}_{\mathcal{DF}}\text{(x)}$ can or can not be equal. The proposed gradient similarity ($\widehat{\text{GS}}$) is computed by:
\begin{equation}
  \ \widehat{\text{GS}} \text{(x)} = \text{GS(x)} + \big[ \text{GS}_{\mathcal{DF}}\text{(x)} - \text{GS}_{\mathcal{RF}}\text{(x)} \big].
  \label{GSprop}
\end{equation}     

The added term $\big[ \text{GS}_{\mathcal{DF}}\text{(x)} - \text{GS}_{\mathcal{RF}}\text{(x)} \big]$, will put more emphasize on removed edges from $\mathcal{R}$ than added edges to the $\mathcal{D}$. For weak added/removed edges, it is likely that weak edges smooth out in $\mathcal{F}$. Therefore, $\big[ \text{GS}_{\mathcal{DF}}\text{(x)} - \text{GS}_{\mathcal{RF}}\text{(x)} \big]$ always put less emphasize on weak edges.    

Comparing visually some outputs of the GS and $\widehat{\text{GS}}$ at this step might not be fair because they have different numerical scales. GS is bounded between 0 and 1, while $\widehat{\text{GS}}$ might have negative values greater than -1, and/or positive values smaller than +2. Therefore, this comparison is performed on the final similarity map and is presented in subsection \ref{GCSanalysis} as well as more explanation on how the proposed $\widehat{\text{GS}}$ works.

\subsection{Chromaticity Similarity}
\label{Chromaticity}

For the case of color changes and especially when the structure of the distorted image remains unchanged, the gradient similarity (GS) and the proposed $\widehat{\text{GS}}$ may lead to inaccurate quality predictions. Therefore, previous researches such as \cite{FSIM, VSI} used a color similarity map to measure color differences. Let $H$ and $M$ denote two chromaticity channels regardless of the type of the color space. In \cite{FSIM, VSI}, for each channel a color similarity is computed and their result is combined as:
\begin{equation}
\small
  \ \text{CS(x)}=\frac{2 H_\mathcal{R}(x) H_\mathcal{D}(x) + C_3} {H_\mathcal{R}^2(x) + H_\mathcal{D}^2(x) + C_3} \times \frac{2 M_\mathcal{R}(x) M_\mathcal{D}(x) + C_3} {M_\mathcal{R}^2(x) + M_\mathcal{D}^2(x) + C_3}
  \label{CS}
\end{equation}                    
where $C_3$ is a constant to control numerical stability. In this paper, we propose a new formulation to calculate color similarity. The proposed formulation calculates a color similarity map using both chromaticity channels at once:    
\begin{equation}
\small
  \ \widehat{\text{CS}} \text{(x)} = \frac{2 \Big( H_\mathcal{R}(x) H_\mathcal{D}(x) + M_\mathcal{R}(x) M_\mathcal{D}(x)\Big) + C_3} {H_\mathcal{R}^2(x) + H_\mathcal{D}^2(x) + M_\mathcal{R}^2(x) + M_\mathcal{D}^2(x) + C_3}
  \label{CSprop}
\end{equation}                    
Similar to the CS in equation (\ref{CS}), the above joint color similarity ($\widehat{\text{CS}}$) formulation gives equal weight to both chromaticity channels $H$ and $M$. It is clear that $\widehat{\text{CS}}$ is more efficient than CS. CS needs 7 multiplications, 6 summations, 2 divisions, and 2 shift operations (multiplications by 2), while $\widehat{\text{CS}}$ needs 6 multiplications, 6 summations, 1 division, and 1 shift operation. Note that CS can also be computed through 8 multiplications, 6 summations, 1 division, and 2 shift operations. In experimental results section, an experiment is conducted to compare usefulness of the CS and $\widehat{\text{CS}}$ along with the proposed metric.   

The gradient similarity maps (GS or $\widehat{\text{GS}}$) can be combined with the joint color similarity map $\widehat{\text{CS}}$ through the following summation (weighted average) scheme:       
 \begin{equation}
  \ \text{GCS(x)}=\alpha \text{GS(x)} + (1-\alpha)\widehat{\text{CS}}\text{(x)}
  \label{GJCS1}
\end{equation} 
\begin{equation}
  \ \widehat{\text{GCS}}\text{(x)}=\alpha \widehat{\text{GS}}\text{(x)} + (1-\alpha)\widehat{\text{CS}}\text{(x)}
  \label{GJCS2}
\end{equation} 
where the parameter $0 \leq \alpha \leq 1$ adjusts the relative importance of the gradient and chromaticity similarity maps. The proposed metric MDSI uses equation (\ref{GJCS2}). Equation (\ref{GJCS1}) is included to be compared with equation (\ref{GJCS2}). An alternative combination scheme which is very popular in state-of-the-art is through multiplication in the form of $[{\widehat{\text{GS}}\text{(x)}}] ^ \gamma  [{\widehat{\text{CS}}\text{(x)}}] ^ \beta$, where the parameters $\gamma$ and $\beta$ are used to adjust the relative importance of the two similarity maps. For several reasons, the proposed index uses the summation scheme (refer to subsection \ref{vs2}).   
        
%

In Fig \ref{fig:out1}, two examples are provided to show that these two similarity maps, e.g. GS and $\widehat{\text{CS}}$, are complementary. In the first example, there is a considerable difference between the gradient maps of the reference and the distorted images. Hence, the GS map is enough for a good judgment. However, this difference in the second example (second row) is trivial, which leads to a wrong prediction by using GS as the only similarity map. The examples in Fig \ref{fig:out1} show that the gradient similarity and chromaticity similarity are complementary.

\subsection{Deviation Pooling}

\begin{figure}[htb]
\scriptsize
\begin{minipage}[b]{0.99\linewidth}
  \centering
  \centerline{\includegraphics[height=4.5cm]{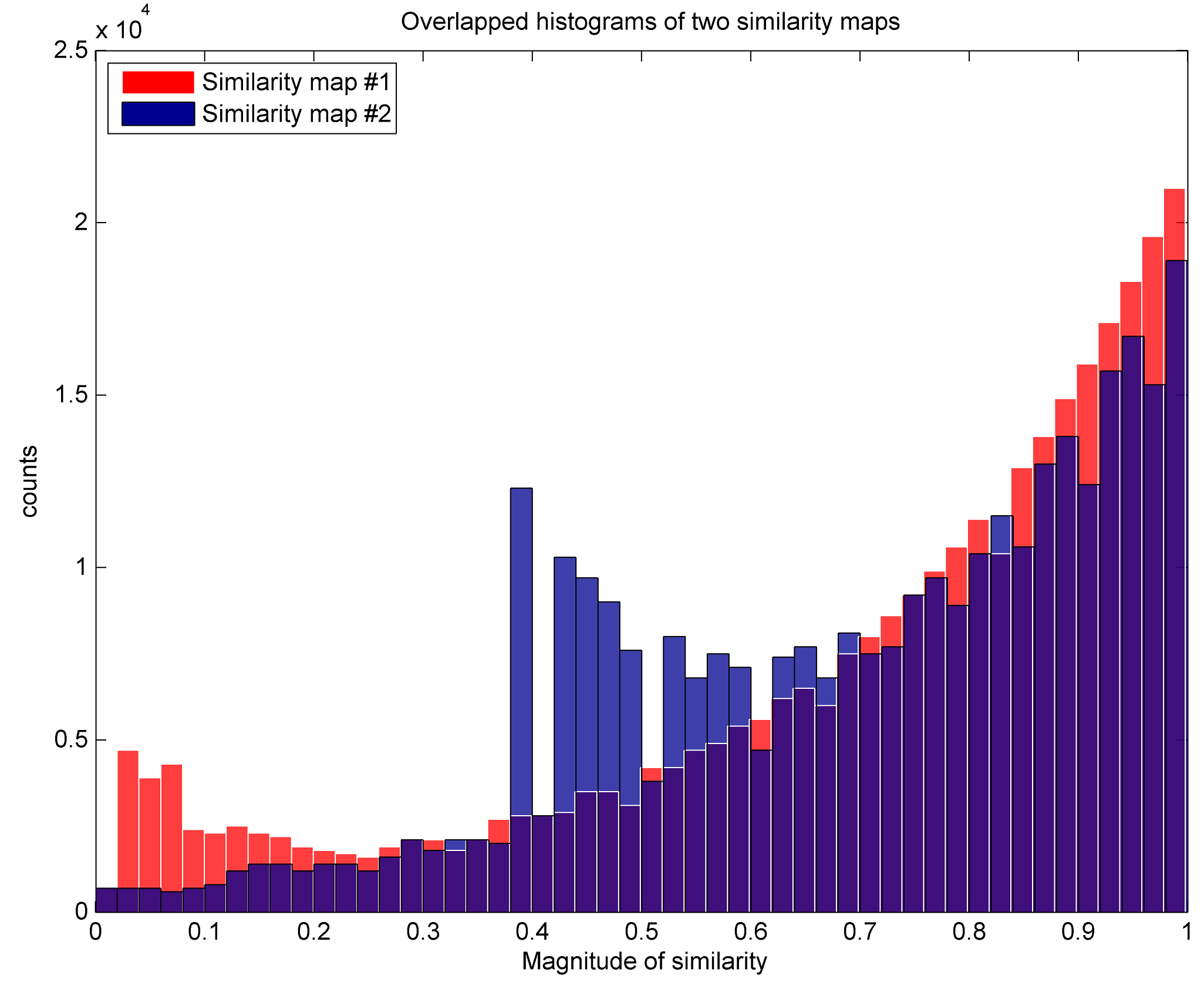}} 
  \vspace{-0.20cm}
\end{minipage}
\caption{Overlapped histograms of two similarity maps corresponding to two distorted images. Lower values of similarity maps indicate to more severe distortions, while higher values refer to less/non distorted pixels.}
\label{MeanMAD}
\end{figure}

The motivation of using the deviation pooling is that HVS is sensitive to both magnitude and the spread of the distortions across the image. Other pooling strategies such as Minkowski pooling and percentile pooling adjust the magnitude of distortions or discard the less/non distorted pixels. These pooling strategies and the mean pooling do not take into account the spread of the distortions. It is shown in \cite{GMSD} by case examples and experimental results that a common wrong prediction by mean pooling is where it calculates the same quality scores for two distorted images of different type. In such cases, deviation pooling is likely to provide good judgments over their quality through spread of the distortions. This is the reason why mean pooling have good inter-class (one distortion type) quality prediction but its performance might be degraded for intra-class (whole dataset) quality prediction. While this statement can be verified from the experimental results provided in \cite{GMSD}, an example is also provided in subsection \ref{vs1} to support this statement. Human visual system penalizes severer distortions much more than the distortion-free regions, and these pixels may constitute different fractions of distorted images. Mean pooling, however, depending on this fraction, is likely to nullify the impact of the severer distortions by inclusion of distortion-free regions into the average computation. Fig. \ref{MeanMAD} shows overlapped histograms of two similarity maps corresponding to two distorted images. While mean pooling indicate that image \#1 is of better quality than image \#2 ($\mu_1>\mu_2$), deviation pooling provides an opposite assessment ($\sigma_1>\sigma_2$). Given that $\mu_1>\mu_2$, and that image \#1 has more severe distortions compared to image \#2 with their values farther from $\mu_1$ than $\mu_2$, there are larger deviations in similarity map of image \#1 than that of image \#2. Therefore, deviation pooling is an alternative to the mean pooling that can also measure different levels of distortions. In the following, we propose the deviation pooling (DP) strategy and provide a general formulation of this pooling. 
DP for IQAs is rarely used in the literature, except the standard deviation used in GMSD \cite{GMSD}, which is a special case of DP. A deviation can be seen as the Minkowski distance of order $\rho$ between vector \textbf{x} and its MCT (Measure of Central Tendency):  
\begin{equation}
  \ \text{DP}^{(\rho)} = \Big( \frac{1}{N} \sum_{i=1}^{N}\big| \textbf{x}_i - \text{MCT} \big|^\rho \Big)^{1/\rho}.
  \label{equ:minkowski1}
\end{equation}                    
where $\rho \geq 1$ indicates the type of deviation. The only MCT that is used in this paper is mean. Though other MCTs such as median and mode can be used, we found that these MCTs do not provide satisfactory quality predictions. 

Several researches shown that more emphasis on the severer distortions can lead to more accurate predictions \cite{spatial2006, percentile2009}. The Minkowski pooling \cite{spatial2006} and the percentile pooling \cite{percentile2009} are two examples. As mentioned before, these pooling strategies follow a property of HVS that penalize severer distortions much more than the less distorted ones even though they constitute a small portion of total distortions. Hence, they try to moderate the weakness of the mean pooling through adjusting magnitudes of distortions \cite{spatial2006} or discarding the less/non distorted regions \cite{percentile2009}. The deviation pooling can be generalized to consider the aforementioned property of HVS:
\begin{equation}
  \ \text{DP}^{(\rho, q)} = \Big( \frac{1}{N} \sum_{i=1}^{N}\big| \textbf{x}_i^q - \text{MCT} \big|^\rho \Big)^{1/\rho}.
  \label{equ:DP}
\end{equation}                    
where, $q$ adjusts the emphasis of the values in vector \textbf{x}, and MCT is calculated through \textbf{x}$_i^q$ values. Furthermore, we propose to use power pooling in conjunction with the deviation pooling to control numerical behavior of the final quality scores:
\begin{equation}
  \ \text{DP}^{(\rho, q, o)} = \bigg[ \Big( \frac{1}{N} \sum_{i=1}^{N}\big| \textbf{x}_i^q - \text{MCT} \big|^\rho \Big)^{1/\rho} \bigg]^o.
  \label{equ:DP}
\end{equation}                    
where, $o$ is the power pooling applied on the final value of the deviation. The power pooling can be used to make an IQA model more linear versus the subjective scores or might be used for better visualization of the scores. Linearity might not be a significant advantage of an IQA, but it is pointed to be of interest in \cite{GMSD}. Also, according to \cite{ITUT2012}, linearity against subjective data is one of the measures for validation of IQAs that should be examined\footnote{Though linearity is measured after a nonlinear analysis.}. The power pooling can also have small impact on the values of Pearson linear Correlation Coefficient (PCC) and Root Mean Square Error (RMSE). Note that the above deviation pooling is equal to the Minkowski pooling \cite{spatial2006} when MCT $=0$, $\rho=1$ and $o=1$. It is equal to the mean absolute deviation (MAD) to the power of $o$ for $\rho=1$, and equal to the standard deviation (SD) to the power of $o$ for $\rho=2$. The three parameters should be set according to the IQA model. More analysis on these three parameters can be found in experimental results section. For the proposed index MDSI, we set $\rho=1$, $q=\frac{1}{4}$ and $o=\frac{1}{4}$. Therefore, the proposed IQA model can be written as:
\begin{equation}
\small
  \ \text{MDSI} = \bigg[ \frac{1}{N} \sum_{i=1}^{N}\big| \widehat{\text{GCS}}_i^{1/4} - \big( \frac{1}{N}\sum_{i=1}^N\widehat{\text{GCS}}_i^{1/4} \big) \big| \bigg]^{1/4}.
  \label{equ:MDSIplus}
\end{equation}                    

\begin{figure*}[htb]
\scriptsize
\begin{minipage}[b]{0.195\linewidth}
  \centering
  \centerline{\includegraphics[height=2.2cm]{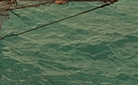}} 
\centerline{$\mathcal{R}$}
\end{minipage}
\begin{minipage}[b]{0.195\linewidth}
  \centering
  \centerline{\includegraphics[height=2.2cm]{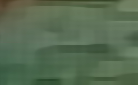}}
\centerline{$\mathcal{D}$}
\end{minipage}
\begin{minipage}[b]{0.195\linewidth}
  \centering
  \centerline{\includegraphics[height=2.2cm]{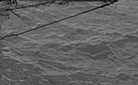}}
\centerline{{$L_\mathcal{R}$}}
\end{minipage}
\begin{minipage}[b]{0.195\linewidth}
  \centering
  \centerline{\includegraphics[height=2.2cm]{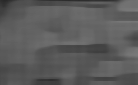}} 
\centerline{{$L_\mathcal{D}$}}
\end{minipage}
\begin{minipage}[b]{0.195\linewidth}
  \centering
  \centerline{\includegraphics[height=2.2cm]{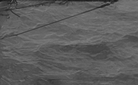}} 
\centerline{{$L_\mathcal{F}$}}
\end{minipage}
\\ \\
\begin{minipage}[b]{0.195\linewidth}
  \centering
  \centerline{\includegraphics[height=2.2cm]{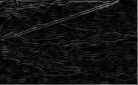}}
  \vspace{0.10cm}
\centerline{$G_\mathcal{R}$}
\end{minipage}
\begin{minipage}[b]{0.195\linewidth}
  \centering
  \centerline{\includegraphics[height=2.2cm]{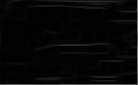}}
  \vspace{0.10cm}
\centerline{$G_\mathcal{D}$}
\end{minipage}
\begin{minipage}[b]{0.195\linewidth}
  \centering
  \centerline{\includegraphics[height=2.2cm]{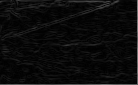}}
  \vspace{0.10cm}
\centerline{$G_\mathcal{F}$}
\end{minipage}
\begin{minipage}[b]{0.195\linewidth}
  \centering
  \centerline{\includegraphics[height=2.2cm]{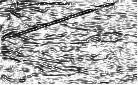}}
  \vspace{0.10cm}
\centerline{$\text{GS}_\mathcal{RD}$ (GS)}
\end{minipage}
\begin{minipage}[b]{0.195\linewidth}
  \centering
  \centerline{\includegraphics[height=2.2cm]{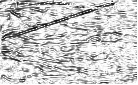}}
  \vspace{0.10cm}
\centerline{$\text{GS}_\mathcal{DF}$}
\end{minipage}
\\ \\
\begin{minipage}[b]{0.195\linewidth}
  \centering
  \centerline{\includegraphics[height=2.2cm]{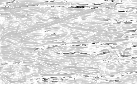}}
  \vspace{0.10cm}
\centerline{$\text{GS}_\mathcal{RF}$}
\end{minipage}
\begin{minipage}[b]{0.195\linewidth}
  \centering
  \centerline{\includegraphics[height=2.2cm]{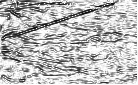}}
  \vspace{0.10cm}
\centerline{GCS}
\end{minipage}
\begin{minipage}[b]{0.195\linewidth}
  \centering
  \centerline{\includegraphics[height=2.2cm]{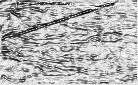}}
  \vspace{0.10cm}
\centerline{$\widehat{\text{GCS}}$}
\end{minipage}
\begin{minipage}[b]{0.195\linewidth}
  \centering
  \centerline{\includegraphics[height=2.2cm]{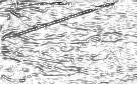}}
  \vspace{0.10cm}
\centerline{GCS$^{1/4}$}
\end{minipage}
\begin{minipage}[b]{0.195\linewidth}
  \centering
  \centerline{\includegraphics[height=2.2cm]{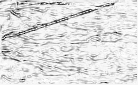}}
  \vspace{0.10cm}
\centerline{$\widehat{\text{GCS}}^{1/4}$}
\end{minipage}
\caption{The difference between similarity maps GCS and $\widehat{\text{GCS}}$ that use conventional gradient similarity and the proposed gradient similarity, respectively.}
\label{GCS14_1}
\end{figure*}

Note that possible interval for $\widehat{\text{GCS}}$ is $[0-\delta_1 ~ 1+\delta_2]$, where $\delta_1<1$ and $\delta_2<1$. It is worth to mention that values of $\widehat{\text{GCS}}$ mostly remain in $[0 ~ 1]$. Also, $\widehat{\text{GCS}} < 0$ are highly distorted pixels, while $\widehat{\text{GCS}} > (1-\epsilon)$ refer to less/non-distorted pixels, where $\epsilon < 1$ is a very small number. The global variations of $\widehat{\text{GCS}}^{1/4}$ is computed by mean absolute deviation, which is followed by power pooling. Note that since absolute of deviations is computed, the quality scores are positive. Larger values of the quality predictions provided by the proposed index indicate to the more severe distorted images, while an image with perfect quality is assessed by a quality score of zero since there is no variation in its similarity map. The important point on the use of the Minkowski pooling on final similarity maps is that terms like ``more emphasize" and ``less emphasize", regardless of the $q$ values have been used, depends also on the pooling strategy and underlying similarity map. For example, placing more emphasize on highly distorted regions by Minkowski pooling will decrease the quality score computed by the mean pooling, but the quality score provided by the deviation pooling might become larger or smaller depending on the spread of the distortions which is directly related to the underlying similarity map.

\subsection{Analysis and Examples of GCS Maps}
\label{GCSanalysis}

In this section, final similarity maps after applying the Minkowski pooling, e.g. GCS$^{1/4}$ and $\widehat{\text{GCS}}^{1/4}$, are compared along with sufficient explanations. The difference between these two similarity map is their use of gradient similarity. GCS uses conventional GS, while $\widehat{\text{GCS}}$ uses the proposed gradient similarity $\widehat{\text{GS}}$. The best way to analyze the effect of the proposed gradient similarity is through step by step explanation and visualization of different examples. In subsection \ref{GSsection}, several disadvantages of the traditional GS was mentioned. Here, each of them are explained and examples are provided.

\paragraph{Case 1 (Removed edge)}  Missing edges in distorted image with respect to its original image means that structural information are removed, hence this disappearance bring attention of the HVS. These regions has to be strongly highlighted in the similarity map.  

\paragraph{Case 2 (A weak added/removed edge)} An extra edge in $\mathcal{D}$ or a removed edge from $\mathcal{R}$
bring less attention of HVS if its color is close to the relative
pixels of that edge in $\mathcal{R}$ ($\mathcal{D}$), or simply it is a weak edge.       

Fig. \ref{GCS14_1} shows how the proposed gradient similarity map $\widehat{\text{GS}}$ performs for \textit{case 1} and \textit{case 2} as a part of the $\widehat{\text{GCS}}$ compared to the GS for GCS. We can see that $\text{GS}_\mathcal{RD}$ (GS) highlighted differences with details. The edges corresponding to the location of ropes in original image are mainly replaced with pixels of another color (dark replaced with green), but many other edges with smaller strengths in $\mathcal{R}$ are replaced with pixels having the same color (green). This latter holds for added edges to the distorted image. In fused image ($L_\mathcal{F}$), some of these weaker edges are smoothed. This can be seen by comparing $\text{GS}_\mathcal{RD}$ and $\text{GS}_\mathcal{DF}$. Both $\text{GS}_\mathcal{RD}$ and $\text{GS}_\mathcal{DF}$ indicate to high differences at the location of the ropes. $\text{GS}_\mathcal{RD} + \text{GS}_\mathcal{DF}$ will also put high emphasize on this location, but less emphasize on the weaker edges. The results is then subtracted by $\text{GS}_\mathcal{RF}$ which in turn again less emphasize is placed on the weak edges (relevant to the darker pixels in $\text{GS}_\mathcal{RF}$). Note that GCS and $\widehat{\text{GCS}}$ have different numerical behavior, so it is fair to compare them by looking at the GCS$^{1/4}$ and $\widehat{\text{GCS}}^{1/4}$. Compared to the GCS$^{1/4}$, $\widehat{\text{GCS}}^{1/4}$ indicate to larger differences at the location of ropes, but smaller differences elsewhere.

    \begin{figure}[htb]
\scriptsize
\begin{minipage}[b]{0.245\linewidth}
  \centering
  \centerline{\includegraphics[height=4cm]{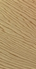}} 
\centerline{$\mathcal{R}$}
\end{minipage}
\begin{minipage}[b]{0.245\linewidth}
  \centering
  \centerline{\includegraphics[height=4cm]{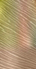}}
\centerline{$\mathcal{D}$}
\end{minipage}
\begin{minipage}[b]{0.245\linewidth}
  \centering
  \centerline{\includegraphics[height=4cm]{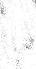}}
\centerline{$\text{GS}_\mathcal{RD}$}
\end{minipage}
\begin{minipage}[b]{0.245\linewidth}
  \centering
  \centerline{\includegraphics[height=4cm]{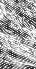}} 
\centerline{$\text{GS}_\mathcal{DF}$}
\end{minipage}
\\ \\
\begin{minipage}[b]{0.245\linewidth}
  \centering
  \centerline{\includegraphics[height=4cm]{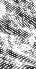}} 
\centerline{$\text{GS}_\mathcal{RF}$}
\end{minipage}
\begin{minipage}[b]{0.245\linewidth}
  \centering
  \centerline{\includegraphics[height=4cm]{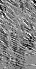}}
\centerline{$(\text{GS}_\mathcal{DF} - \text{GS}_\mathcal{RF})$}
\end{minipage}
\begin{minipage}[b]{0.245\linewidth}
  \centering
  \centerline{\includegraphics[height=4cm]{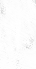}}
\centerline{GCS$^{1/4}$}
\end{minipage}
\begin{minipage}[b]{0.245\linewidth}
  \centering
  \centerline{\includegraphics[height=4cm]{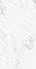}} 
\centerline{$\widehat{\text{GCS}}^{1/4}$}
\end{minipage}
\caption{The difference between similarity maps GCS$^{1/4}$ and $\widehat{\text{GCS}}^{1/4}$ for the case of the inverted edges. Note that some intermediate outputs are not shown.}
\label{coloredge}
\end{figure}

\paragraph{Case 3 (Preserved edge but with different color)} Although a color similarity map should measure color differences at the location of the inverted edges, edges constitute a small fraction of total pixels in images, and it is common to give smaller weights to a color similarity map than structural similarities such as gradient similarity. While traditional gradient similarity does not work well in this situation, the proposed gradient similarity can partially solve this problem. Fig. \ref{coloredge} provides an example in which most of the edges are inverted in the distorted image. We can see that $\widehat{\text{GCS}}^{1/4}$ highlighted much more differences than GCS$^{1/4}$ at these locations, thanks to the added term $(\text{GS}_\mathcal{DF} - \text{GS}_\mathcal{RF})$ to the traditional gradient similarity. In fact, $\text{GS}_\mathcal{DF}$ is likely to be different than $\text{GS}_\mathcal{RF}$ in this case because these edges in $\mathcal{F}$ are likely to become closer to their surrounding pixels in either $\mathcal{R}$ or $\mathcal{D}$ images.

\section{Experimental results and discussion}
\label{results}

\begin{figure}[htb]
\scriptsize
\begin{minipage}[b]{0.99\linewidth}
  \centering
  \centerline{\includegraphics[height=2.8cm]{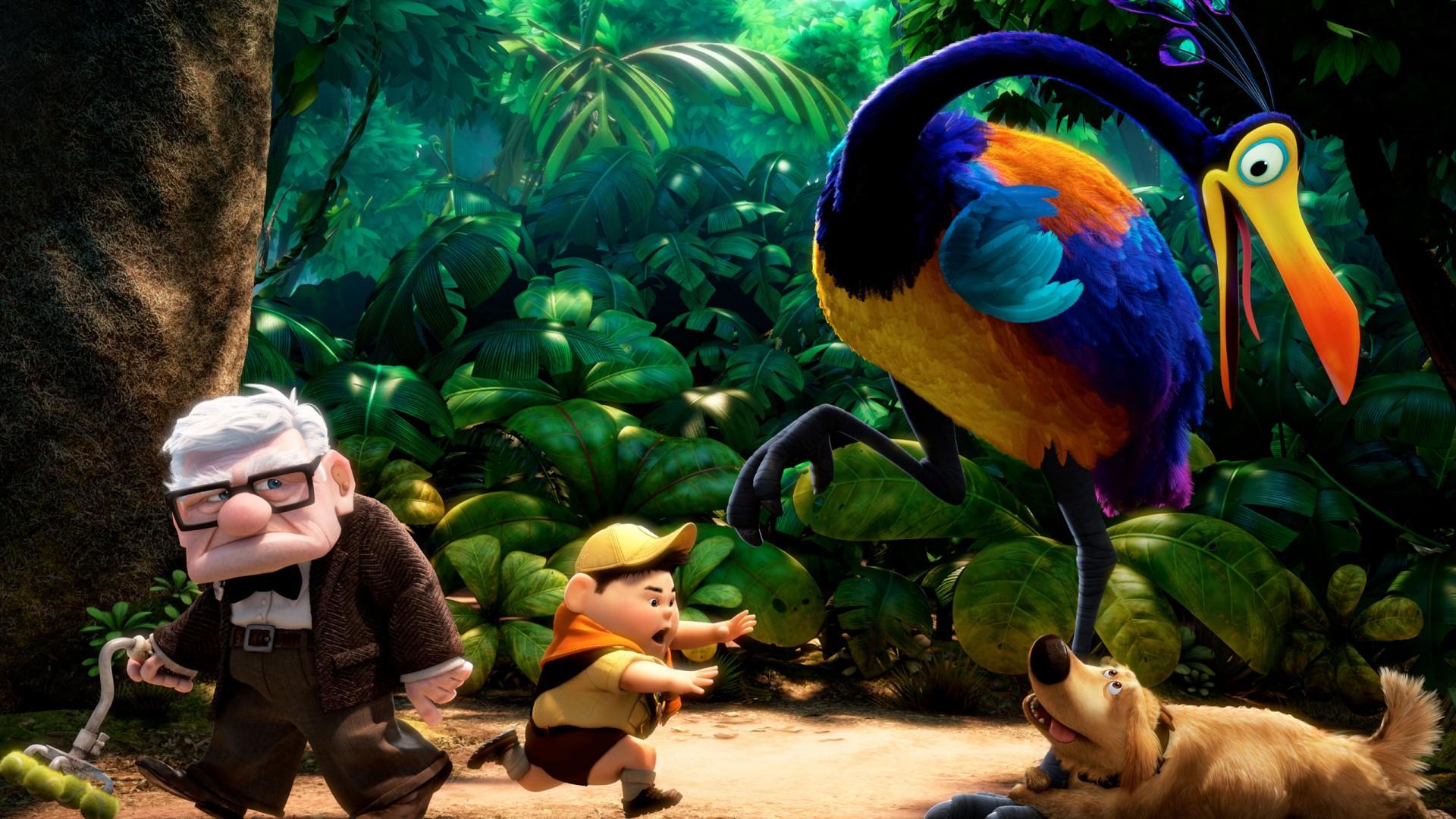}} 
  \vspace{0.10cm}
\centerline{$\mathcal{R}$}
\end{minipage}
\\ \\
\begin{minipage}[b]{.99\linewidth}
  \centering
  \centerline{\includegraphics[height=2.8cm]{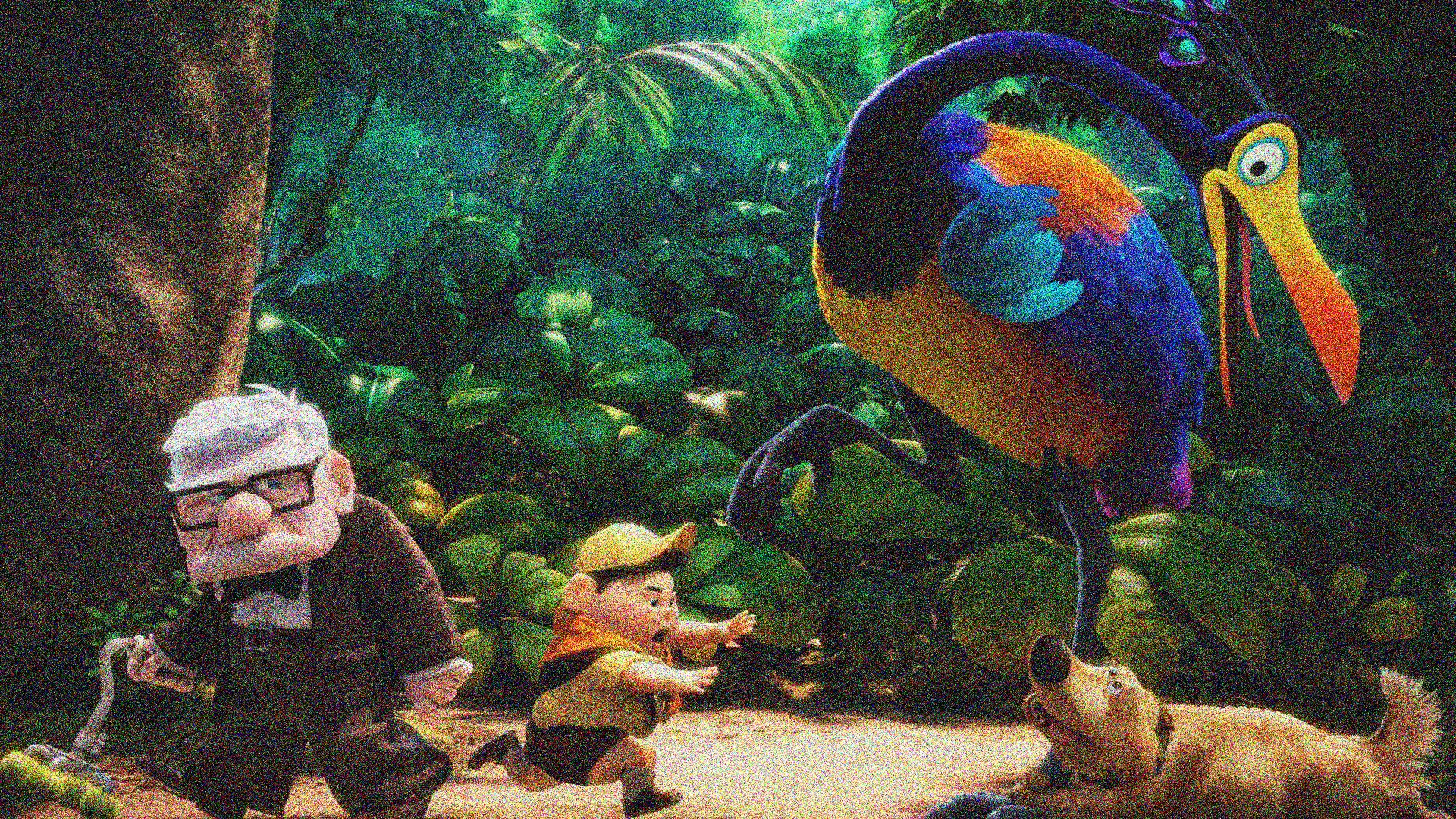}}
  \vspace{0.10cm}
\centerline{$\mathcal{D}$ (Gaussian noise)}
\end{minipage}
\caption{An example of reference $\mathcal{R}$ and distorted $\mathcal{D}$ image in the ESPL synthetic images database \cite{ESPL}.}
\label{fig:synthetic}
\end{figure}

In the experiments, eight datasets were used. The LIVE dataset \cite{LIVEweb} contains 29 reference images and 779 distorted images of five categories. The TID2008 \cite{TID2008} dataset contains 25 reference images and 1700 distorted images. For each reference image, 17 types of distortions of 4 degrees are available. CSIQ \cite{MAD} is another dataset that consists of 30 reference images; each is distorted using six different types of distortions at four to five levels of distortion. The large TID2013 \cite{TID2013} dataset contains 25 reference images and 3000 distorted images. For each reference image, 24 types of distortions of 5 degrees are available. VCL@FER database \cite{VCL} consists of 23 reference images and 552 distorted images, with four degradation types and six degrees of degradation. In addition to these five datasets, contrast distorted images of the CCID2014 dataset \cite{CCID2014} are used in the experiments. This dataset contains 655 contrast distorted images of five types. Gamma transfer, convex and concave arcs, cubic and logistic functions, mean shifting, and a compound function are used to generate these five types of distortions. We also used the ESPL synthetic image database \cite{ESPL} which contains 25 synthetic images of video games and animated movies. It contains 500 distorted images of 5 categories. Fig. \ref{fig:synthetic} shows an example of a reference and a distorted synthetic image. Finally, the digitally retouched image quality (DRIQ) dataset \cite{DRIQ} was used in the experiments. It contains 26 reference images and 3 enhanced images for each reference image.

\begin{table*}[htb]
\scriptsize
\centering
\caption{Performance comparison of the proposed IQA model, MDSI, and twelve popular/competing indices on eight benchmark datasets. Note that top three IQA models are highlighted.}
\begin{tabular}{ccccccccccccccc}
\hline
                                                                         &      & MSSSIM & VIF             & MAD             & IWSSIM & SR\_SIM         & FSIM$_c$           & GMSD            & SFF             & VSI             & DSCSI           & ADD-GSIM        & SCQI            & MDSI            \\ \hline
                                                                         & SRC  & 0.8542 & 0.7491          & 0.8340          & 0.8559 & 0.8913          & 0.8840          & 0.8907          & 0.8767          & 0.8979          & 0.8634          & \textbf{0.9094} & \textbf{0.9051} & \textbf{0.9208} \\
TID                                                                      & PCC  & 0.8451 & 0.8084          & 0.8290          & 0.8579 & 0.8866          & 0.8762          & 0.8788          & 0.8817          & 0.8762          & 0.8445          & \textbf{0.9120} & \textbf{0.8899} & \textbf{0.9160} \\
2008                                                                     & KRC  & 0.6568 & 0.5861          & 0.6445          & 0.6636 & 0.7149          & 0.6991          & 0.7092          & 0.6882          & 0.7123          & 0.6651          & \textbf{0.7389} & \textbf{0.7294} & \textbf{0.7515} \\
                                                                         & RMSE & 0.7173 & 0.7899          & 0.7505          & 0.6895 & 0.6206          & 0.6468          & 0.6404          & 0.6333          & 0.6466          & 0.7187          & \textbf{0.5504} & \textbf{0.6120} & \textbf{0.5383} \\ \hline
                                                                         & SRC  & 0.9133 & 0.9195          & 0.9467          & 0.9213 & 0.9319          & 0.9310          & \textbf{0.9570} & \textbf{0.9627} & 0.9423          & 0.9417          & 0.9422          & 0.9434          & \textbf{0.9569} \\
\multirow{2}{*}{CSIQ}                                                    & PCC  & 0.8991 & 0.9277          & 0.9500          & 0.9144 & 0.9250          & 0.9192          & \textbf{0.9541} & \textbf{0.9643} & 0.9279          & 0.9313          & 0.9342          & 0.9268          & \textbf{0.9531} \\
                                                                         & KRC  & 0.7393 & 0.7537          & 0.7970          & 0.7529 & 0.7725          & 0.7690          & \textbf{0.8129} & \textbf{0.8288} & 0.7857          & 0.7787          & 0.7894          & 0.7870          & \textbf{0.8130} \\
                                                                         & RMSE & 0.1149 & 0.0980          & 0.0820          & 0.1063 & 0.0997          & 0.1034          & \textbf{0.0786} & \textbf{0.0695} & 0.0979          & 0.0956          & 0.0937          & 0.0986          & \textbf{0.0795} \\ \hline
                                                                         & SRC  & 0.9513 & 0.9636          & \textbf{0.9669} & 0.9567 & 0.9618          & 0.9645          & 0.9603          & 0.9649          & 0.9524          & 0.9487          & \textbf{0.9681} & 0.9406          & \textbf{0.9667} \\
\multirow{2}{*}{LIVE}                                                    & PCC  & 0.9489 & 0.9604          & \textbf{0.9675} & 0.9522 & 0.9553          & 0.9613          & 0.9603          & 0.9632          & 0.9482          & 0.9434          & \textbf{0.9657} & 0.9344          & \textbf{0.9659} \\
                                                                         & KRC  & 0.8044 & 0.8282          & \textbf{0.8421} & 0.8175 & 0.8299          & 0.8363          & 0.8268          & 0.8365          & 0.8058          & 0.7982          & \textbf{0.8474} & 0.7835          & \textbf{0.8395} \\
                                                                         & RMSE & 8.6188 & 7.6137          & \textbf{6.9072} & 8.3472 & 8.0812          & 7.5296          & 7.6214          & 7.3460          & 8.6817          & 9.0635          & \textbf{7.0925} & 9.7355          & \textbf{7.0790} \\ \hline
                                                                         & SRC  & 0.7859 & 0.6769          & 0.7807          & 0.7779 & 0.8073          & 0.8510          & 0.8044          & 0.8513          & \textbf{0.8965} & 0.8744          & 0.8285          & \textbf{0.9052} & \textbf{0.8899} \\
TID                                                                      & PCC  & 0.8329 & 0.7720          & 0.8267          & 0.8319 & 0.8663          & 0.8769          & 0.8590          & 0.8706          & \textbf{0.9000} & 0.8782          & 0.8807          & \textbf{0.9071} & \textbf{0.9085} \\
2013                                                                     & KRC  & 0.6047 & 0.5147          & 0.6035          & 0.5977 & 0.6406          & 0.6665          & 0.6339          & 0.6581          & \textbf{0.7183} & 0.6862          & 0.6646          & \textbf{0.7327} & \textbf{0.7123} \\
                                                                         & RMSE & 0.6861 & 0.7880          & 0.6976          & 0.6880 & 0.6193          & 0.5959          & 0.6346          & 0.6099          & \textbf{0.5404} & 0.5930          & 0.5871          & \textbf{0.5219} & \textbf{0.5181} \\ \hline
\multicolumn{1}{l}{}                                                     & SRC  & 0.9227 & 0.8866          & 0.9061          & 0.9163 & 0.9021          & \textbf{0.9323} & 0.9177          & 0.7738          & 0.9317          & 0.9289          & \textbf{0.9366} & 0.9083          & \textbf{0.9318} \\
VCL@                                                                     & PCC  & 0.9232 & 0.8938          & 0.9053          & 0.9191 & 0.9023          & 0.9329          & 0.9176          & 0.7761          & 0.9320          & \textbf{0.9338} & \textbf{0.9339} & 0.9107          & \textbf{0.9349} \\
FER                                                                      & KRC  & 0.7497 & 0.6924          & 0.7213          & 0.7372 & 0.7183          & \textbf{0.7643} & 0.7406          & 0.5779          & \textbf{0.7633}          & 0.7588          & \textbf{0.7731} & 0.7316          & 0.7629 \\
\multicolumn{1}{l}{}                                                     & RMSE & 9.4398 & 11.014          & 10.433          & 9.6788 & 10.589          & 8.8480          & 9.7643          & 15.488          & 8.9051          & \textbf{8.7902} & \textbf{8.7819} & 10.147          & \textbf{8.7136} \\ \hline
                                                                         & SRC  & 0.7770 & \textbf{0.8349} & 0.7451          & 0.7811 & 0.7363          & 0.7657          & 0.8077          & 0.6859          & 0.7734          & 0.7400          & \textbf{0.8698} & 0.7811          & \textbf{0.8128} \\
CCID                                                                     & PCC  & 0.8278 & \textbf{0.8588} & 0.7516          & 0.8342 & 0.7834          & 0.8204          & 0.8521          & 0.7575          & 0.8209          & 0.7586          & \textbf{0.8935} & 0.8200          & \textbf{0.8576} \\
2014                                                                     & KRC  & 0.5845 & \textbf{0.6419} & 0.5490          & 0.5898 & 0.5372          & 0.5707          & 0.6100          & 0.5012          & 0.5735          & 0.5468          & \textbf{0.6840} & 0.5812          & \textbf{0.6181} \\
                                                                         & RMSE & 0.3668 & \textbf{0.3350} & 0.4313          & 0.3606 & 0.4064          & 0.3739          & 0.3422          & 0.4269          & 0.3734          & 0.4260          & \textbf{0.2936} & 0.3734          & \textbf{0.3363} \\ \hline
                                                                         & SRC  & 0.7247 & 0.7488          & 0.8624          & 0.8270 & \textbf{0.8802} & \textbf{0.8766} & 0.8209          & 0.8127          & 0.8717          & 0.7263          & 0.7828          & 0.8292          & \textbf{0.8806} \\
\multirow{2}{*}{ESPL}                                                    & PCC  & 0.7322 & 0.7423          & 0.8677          & 0.8300 & \textbf{0.8732} & \textbf{0.8738} & 0.8234          & 0.8179          & 0.8726          & 0.7302          & 0.7902          & 0.8356          & \textbf{0.8802} \\
                                                                         & KRC  & 0.5208 & 0.5565          & 0.6720          & 0.6221 & \textbf{0.6932} & \textbf{0.6853} & 0.6178          & 0.6127          & 0.6765          & 0.5222          & 0.5814          & 0.6243          & \textbf{0.6895} \\
                                                                         & RMSE & 9.4519 & 9.2985          & 6.8985          & 7.7404 & \textbf{6.7646} & \textbf{6.7482} & 7.8753          & 7.9844          & 6.7791          & 9.4815          & 8.5053          & 7.6241          & \textbf{6.5862} \\ \hline
\multicolumn{1}{l}{}                                                     & SRC  & 0.6692 & 0.8078          & 0.6867          & 0.6903 & 0.7551          & 0.7751          & 0.7762          & \textbf{0.8342} & 0.8222          & 0.8167          & 0.7661          & \textbf{0.8482} & \textbf{0.8508} \\
\multirow{2}{*}{DRIQ}                                                    & PCC  & 0.7058 & \textbf{0.8496} & 0.6967          & 0.7155 & 0.8027          & 0.7989          & 0.8001          & 0.8420          & 0.8477          & 0.8463          & 0.8053          & \textbf{0.8638} & \textbf{0.8702} \\
                                                                         & KRC  & 0.4739 & 0.5997          & 0.4898          & 0.4952 & 0.5604          & 0.5771          & 0.5758          & \textbf{0.6477} & 0.6177          & 0.6104          & 0.5618          & \textbf{0.6490} & \textbf{0.6557} \\
\multicolumn{1}{l}{}                                                     & RMSE & 1.4450 & \textbf{1.0759} & 1.4631          & 1.4249 & 1.2165          & 1.2268          & 1.2235          & 1.1004          & 1.0820          & 1.0864          & 1.2092          & \textbf{1.0277} & \textbf{1.0050} \\ \hline
\multirow{3}{*}{\begin{tabular}[c]{@{}c@{}}Direct\\ Avg.\end{tabular}}   & SRC  & 0.8248 & 0.8234          & 0.8411          & 0.8408 & 0.8583          & 0.8725          & 0.8669          & 0.8453          & \textbf{0.8860} & 0.8550          & 0.8754          & \textbf{0.8826}          & \textbf{0.9013} \\
                                                                         & PCC  & 0.8394 & 0.8516          & 0.8493          & 0.8569 & 0.8743          & 0.8824          & 0.8807          & 0.8591          & \textbf{0.8907} & 0.8583          & \textbf{0.8894}          & 0.8860          & \textbf{0.9108} \\
                                                                         & KRC  & 0.6418 & 0.6466          & 0.6649          & 0.6595 & 0.6834          & 0.6960          & 0.6909          & 0.6689          & \textbf{0.7067} & 0.6708          & \textbf{0.7051}          & 0.7023          & \textbf{0.7303} \\ \hline
\multirow{3}{*}{\begin{tabular}[c]{@{}c@{}}Weighted\\ Avg.\end{tabular}} & SRC  & 0.8335 & 0.7783          & 0.8374          & 0.8387 & 0.8578          & 0.8769          & 0.8626          & 0.8585          & \textbf{0.8974} & 0.8698          & 0.8783          & \textbf{0.8977} & \textbf{0.9066} \\
                                                                         & PCC  & 0.8521 & 0.8287          & 0.8546          & 0.8626 & 0.8810          & 0.8877          & 0.8838          & 0.8729          & 0.8963          & 0.8679          & \textbf{0.8995} & \textbf{0.8967} & \textbf{0.9160} \\
                                                                         & KRC  & 0.6511 & 0.6112          & 0.6626          & 0.6587 & 0.6880          & 0.6999          & 0.6913          & 0.6791          & \textbf{0.7206} & 0.6855          & 0.7140          & \textbf{0.7231} & \textbf{0.7375} \\ \hline
\end{tabular}
\label{results1}
\end{table*}

For objective evaluation, four popular evaluation metrics were used in the experiments: the Spearman Rank-order Correlation coefficient (SRC), the Pearson linear Correlation Coefficient (PCC) after a nonlinear regression analysis (equation \ref{equ:REG}), the Kendall Rank Correlation coefficient (KRC) and the Root Mean Square Error (RMSE). The SRC, PCC, and RMSE metrics measure prediction monotonicity, prediction linearity, and prediction accuracy, respectively. The KRC was used to evaluate the degree of similarity between quality scores and MOS. In addition, Pearson linear Correlation Coefficient \textit{without} nonlinear analysis is used and denoted by LPCC.

Twelve state-of-the-art IQA models were chosen for comparison \cite{MSSSIM, VIF, MAD, IWSSIM, SRSIM, FSIM, GMSD, SFF, VSI, DSCSI, ADD-GSIM, SCQI} including the most recent indices in literature \cite{SFF, GMSD, VSI, DSCSI, ADD-GSIM, SCQI}. It should be noted that the five indices SFF \cite{SFF}, GMSD \cite{GMSD}, VSI \cite{VSI}, \cite{ADD-GSIM}, and SCQI \cite{SCQI} have shown superior performance over state-of-the-art indices.

\subsection{Performance comparison}
\label{performance}

In Table \ref{results1}, the overall performance of thirteen IQA models on eight benchmark datasets, e.g. TID2008, CSIQ, LIVE, TID2013, VCL@FER, CCID2014, ESPL, and DRIQ, is listed. For each dataset and evaluation metric, the top three IQA models are highlighted. On eight datasets, MDSI is 31 times among the top indices, followed by ADD-GSIM (16 times), SCQI (12 times), SFF/FSIM$_c$/VIF (6 times), VSI (5 times), GMSD/SR\_SIM/MAD\footnote{Note the conflict between `MAD' \cite{MAD} as an IQA model, and `MAD' as a pooling strategy.} (4 times), DSCSI (2 times), and MSSSIM/IWSSIM (0 times). To provide a conclusion on the overall performance of these indices, direct and weighted\footnote{The dataset size-weighted average is commonly used in the literature \cite{IWSSIM, SFF, GMSD, VSI}.} overall performances on the eight datasets (8150 images) are also listed in Table \ref{results1}. It can be seen that MDSI has the best overall performance on the eight datasets, while metrics VSI and SCQI are the second, and third best, respectively.

\subsection{Visualization and statistical evaluation}
\label{significance}

\begin{table*}[htb]
\centering{
\tiny
\caption{The results of statistical significance test for ten IQA models on eight datasets. The result of the F-test is equal to +1 if a metric is significantly better than another metric, it is equal to -1 if that metric is statistically inferior to another metric, and the result is equal to 0 if two metrics are statistically indistinguishable. The cumulative sum of individual tests for each metric is listed in the last column with top three IQA models being highlighted in the same column.}
\label{Ftest}
\begin{tabular}{|l|l|c|c|c|c|c|c|c|c|c|c|c|}
\hline
\multicolumn{2}{|c|}{$\rhd$TID2008} & 1           & 2           & 3           & 4           & 5           & 6           & 7           & 8           & 9           & 10 & sum         \\ \hline
1             & VIF                 & -           & -1          & -1          & -1          & -1          & -1          & -1          & -1          & -1          & -1 & -9          \\ \hline
2             & MAD                 & \textbf{+1} & -           & -1          & -1          & -1          & -1          & -1          & -1          & -1          & -1 & -7          \\ \hline
3             & SR\_SIM             & \textbf{+1} & \textbf{+1} & -           & \textbf{+1} & \textbf{+1} & \textbf{+1} & \textbf{+1} & -1          & -1          & -1 & +3          \\ \hline
4             & FSIM$_c$            & \textbf{+1} & \textbf{+1} & -1          & -           & -1          & -1          & \textbf{+1} & -1          & -1          & -1 & -3          \\ \hline
5             & GMSD                & \textbf{+1} & \textbf{+1} & -1          & \textbf{+1} & -           & -1          & \textbf{+1} & -1          & -1          & -1 & -1          \\ \hline
6             & SFF                 & \textbf{+1} & \textbf{+1} & -1          & \textbf{+1} & \textbf{+1} & -           & \textbf{+1} & -1          & -1          & -1 & +1          \\ \hline
7             & VSI                 & \textbf{+1} & \textbf{+1} & -1          & 0           & -1          & -1          & -           & -1          & -1          & -1 & -4          \\ \hline
8             & ADD-GSIM            & \textbf{+1} & \textbf{+1} & \textbf{+1} & \textbf{+1} & \textbf{+1} & \textbf{+1} & \textbf{+1} & -           & \textbf{+1} & -1 & \textbf{+7} \\ \hline
9             & SCQI                & \textbf{+1} & \textbf{+1} & \textbf{+1} & \textbf{+1} & \textbf{+1} & \textbf{+1} & \textbf{+1} & -1          & -           & -1 & \textbf{+5} \\ \hline
10            & MDSI                & \textbf{+1} & \textbf{+1} & \textbf{+1} & \textbf{+1} & \textbf{+1} & \textbf{+1} & \textbf{+1} & \textbf{+1} & \textbf{+1} & -  & \textbf{+9} \\ \hline
\end{tabular}
\hspace*{2 mm}
\begin{tabular}{|l|l|c|c|c|c|c|c|c|c|c|c|c|}
\hline
\multicolumn{2}{|c|}{$\rhd$CSIQ} & 1           & 2           & 3           & 4           & 5           & 6          & 7           & 8           & 9           & 10          & sum         \\ \hline
1            & VIF               & \textbf{-}  & -1          & \textbf{+1} & \textbf{+1} & -1          & -1         & 0           & -1          & 0           & -1          & -3          \\ \hline
2            & MAD               & +1          & \textbf{-}  & \textbf{+1} & \textbf{+1} & -1          & -1         & \textbf{+1} & \textbf{+1} & \textbf{+1} & -1          & +3          \\ \hline
3            & SR\_SIM           & -1          & -1          & \textbf{-}  & \textbf{+1} & -1          & -1         & -1          & -1          & -1          & -1          & -7 \\ \hline
4            & FSIM$_c$          & -1          & -1          & -1          & \textbf{-}  & -1          & -1         & -1          & -1          & -1          & -1          & -9          \\ \hline
5            & GMSD              & \textbf{+1} & \textbf{+1} & \textbf{+1} & \textbf{+1} & \textbf{-}  & -1         & \textbf{+1} & \textbf{+1} & \textbf{+1} & 0           & \textbf{+6} \\ \hline
6            & SFF               & \textbf{+1} & \textbf{+1} & \textbf{+1} & \textbf{+1} & \textbf{+1} & \textbf{-} & \textbf{+1} & \textbf{+1} & \textbf{+1} & \textbf{+1} & \textbf{+9} \\ \hline
7            & VSI               & 0           & -1          & \textbf{+1} & \textbf{+1} & -1          & -1         & \textbf{-}  & -1          & 0           & -1          & -3          \\ \hline
8            & ADD-GSIM          & \textbf{+1} & -1          & \textbf{+1} & \textbf{+1} & -1          & -1         & \textbf{+1} & \textbf{-}  & \textbf{+1} & -1          & +1          \\ \hline
9            & SCQI              & 0           & -1          & \textbf{+1} & \textbf{+1} & -1          & -1         & 0           & -1          & \textbf{-}  & -1          & -3          \\ \hline
10           & MDSI              & \textbf{+1} & \textbf{+1} & \textbf{+1} & \textbf{+1} & 0           & -1         & \textbf{+1} & \textbf{+1} & \textbf{+1} & \textbf{-}  & \textbf{+6} \\ \hline
\end{tabular}
\newline
\vspace*{2 mm}
\newline
\begin{tabular}{|l|l|c|c|c|c|c|c|c|c|c|c|c|}
\hline
\multicolumn{2}{|c|}{$\rhd$LIVE} & 1           & 2          & 3           & 4           & 5           & 6           & 7           & 8          & 9           & 10         & sum         \\ \hline
1            & VIF               & \textbf{-}  & -1         & \textbf{+1} & 0           & 0           & -1          & \textbf{+1} & -1         & \textbf{+1} & -1         & -1          \\ \hline
2            & MAD               & \textbf{+1} & \textbf{-} & \textbf{+1} & \textbf{+1} & \textbf{+1} & \textbf{+1} & \textbf{+1} & 0          & \textbf{+1} & 0          & \textbf{+7} \\ \hline
3            & SR\_SIM           & -1          & -1         & \textbf{-}  & -1          & -1          & -1          & \textbf{+1} & -1         & \textbf{+1} & -1         & -5          \\ \hline
4            & FSIM$_c$          & 0           & -1         & \textbf{+1} & \textbf{-}  & 0           & 0           & \textbf{+1} & -1         & \textbf{+1} & -1         & 0           \\ \hline
5            & GMSD              & 0           & -1         & \textbf{+1} & 0           & \textbf{-}  & -1          & \textbf{+1} & -1         & \textbf{+1} & -1         & -1          \\ \hline
6            & SFF               & \textbf{+1} & -1         & \textbf{+1} & 0           & \textbf{+1} & \textbf{-}  & \textbf{+1} & -1         & \textbf{+1} & -1         & +2          \\ \hline
7            & VSI               & -1          & -1         & -1          & -1          & -1          & -1          & \textbf{-}  & -1         & \textbf{+1} & -1         & -7          \\ \hline
8            & ADD-GSIM          & \textbf{+1} & 0          & \textbf{+1} & \textbf{+1} & \textbf{+1} & \textbf{+1} & \textbf{+1} & \textbf{-} & \textbf{+1} & 0          & \textbf{+7} \\ \hline
9            & SCQI              & -1          & -1         & -1          & -1          & -1          & -1          & -1          & -1         & \textbf{-}  & -1         & -9          \\ \hline
10           & MDSI              & \textbf{+1} & 0          & \textbf{+1} & \textbf{+1} & \textbf{+1} & \textbf{+1} & \textbf{+1} & 0          & \textbf{+1} & \textbf{-} & \textbf{+7} \\ \hline
\end{tabular}
\hspace*{2 mm}
\begin{tabular}{|l|l|c|c|c|c|c|c|c|c|c|c|c|}
\hline
\multicolumn{2}{|c|}{$\rhd$TID2013} & 1           & 2           & 3           & 4           & 5           & 6           & 7           & 8           & 9           & 10 & sum         \\ \hline
1             & VIF                 & \textbf{-}  & -1          & -1          & -1          & -1          & -1          & -1          & -1          & -1          & -1 & -9          \\ \hline
2             & MAD                 & \textbf{+1} & \textbf{-}  & -1          & -1          & -1          & -1          & -1          & -1          & -1          & -1 & -7          \\ \hline
3             & SR\_SIM             & \textbf{+1} & \textbf{+1} & \textbf{-}  & -1          & \textbf{+1} & -1          & -1          & -1          & -1          & -1 & -3          \\ \hline
4             & FSIM$_c$            & \textbf{+1} & \textbf{+1} & \textbf{+1} & \textbf{-}  & \textbf{+1} & \textbf{+1} & -1          & -1          & -1          & -1 & +1          \\ \hline
5             & GMSD                & \textbf{+1} & \textbf{+1} & -1          & -1          & \textbf{-}  & -1          & -1          & -1          & -1          & -1 & -5          \\ \hline
6             & SFF                 & \textbf{+1} & \textbf{+1} & \textbf{+1} & -1          & \textbf{+1} & \textbf{-}  & -1          & 0           & -1          & -1 & 0           \\ \hline
7             & VSI                 & \textbf{+1} & \textbf{+1} & \textbf{+1} & \textbf{+1} & \textbf{+1} & \textbf{+1} & \textbf{-}  & \textbf{+1} & -1          & -1 & \textbf{+5} \\ \hline
8             & ADD-GSIM            & \textbf{+1} & \textbf{+1} & \textbf{+1} & \textbf{+1} & \textbf{+1} & \textbf{+1} & -1          & -           & -1          & -1 & +3          \\ \hline
9             & SCQI                & \textbf{+1} & \textbf{+1} & \textbf{+1} & \textbf{+1} & \textbf{+1} & \textbf{+1} & \textbf{+1} & \textbf{+1} & -           & -1 & \textbf{+7} \\ \hline
10            & MDSI                & \textbf{+1} & \textbf{+1} & \textbf{+1} & \textbf{+1} & \textbf{+1} & \textbf{+1} & \textbf{+1} & \textbf{+1} & \textbf{+1} & -  & \textbf{+9} \\ \hline
\end{tabular}
\newline
\vspace*{2 mm}
\newline
\hspace*{-2 mm}
\begin{tabular}{|l|l|c|c|c|c|c|c|c|c|c|c|c|}
\hline
\multicolumn{2}{|c|}{\multirow{2}{*}{\begin{tabular}[c]{@{}c@{}}$\rhd$VCL\\ ~@FER\end{tabular}}} & \multirow{2}{*}{1} & \multirow{2}{*}{2} & \multirow{2}{*}{3} & \multirow{2}{*}{4} & \multirow{2}{*}{5} & \multirow{2}{*}{6} & \multirow{2}{*}{7} & \multirow{2}{*}{8} & \multirow{2}{*}{9} & \multirow{2}{*}{10} & \multirow{2}{*}{sum} \\
\multicolumn{2}{|c|}{}                                                                          &                    &                    &                    &                    &                    &                    &                    &                    &                    &                     &                      \\ \hline
1                                           & VIF                                               & \textbf{-}         & -1                 & -1                 & -1                 & -1                 & \textbf{+1}        & -1                 & -1                 & -1                 & -1                  & -7                   \\ \hline
2                                           & MAD                                               & \textbf{+1}        & \textbf{-}         & \textbf{+1}        & -1                 & -1                 & \textbf{+1}        & -1                 & -1                 & -1                 & -1                  & -3                   \\ \hline
3                                           & SR\_SIM                                           & \textbf{+1}        & -1                 & \textbf{-}         & -1                 & -1                 & \textbf{+1}        & -1                 & -1                 & -1                 & -1                  & -5                   \\ \hline
4                                           & FSIM$_c$                                          & \textbf{+1}        & \textbf{+1}        & \textbf{+1}        & \textbf{-}         & \textbf{+1}        & \textbf{+1}        & 0                  & 0                  & \textbf{+1}        & 0                   & \textbf{+6}          \\ \hline
5                                           & GMSD                                              & \textbf{+1}        & \textbf{+1}        & \textbf{+1}        & -1                 & \textbf{-}         & \textbf{+1}        & -1                 & -1                 & \textbf{+1}        & -1                  & +1                   \\ \hline
6                                           & SFF                                               & -1                 & -1                 & -1                 & -1                 & -1                 & \textbf{-}         & -1                 & -1                 & -1                 & -1                  & -9                   \\ \hline
7                                           & VSI                                               & \textbf{+1}        & \textbf{+1}        & \textbf{+1}        & 0                  & \textbf{+1}        & \textbf{+1}        & \textbf{-}         & 0                  & \textbf{+1}        & -1                  & +5                   \\ \hline
8                                           & ADD-GSIM                                          & \textbf{+1}        & \textbf{+1}        & \textbf{+1}        & 0                  & \textbf{+1}        & \textbf{+1}        & 0                  & \textbf{-}         & \textbf{+1}        & 0                   & \textbf{+6}          \\ \hline
9                                           & SCQI                                              & \textbf{+1}        & \textbf{+1}        & \textbf{+1}        & -1                 & -1                 & \textbf{+1}        & -1                 & -1                 & \textbf{-}         & -1                  & -1                   \\ \hline
10                                          & MDSI                                              & \textbf{+1}        & \textbf{+1}        & \textbf{+1}        & 0                  & \textbf{+1}        & \textbf{+1}        & \textbf{+1}        & 0                  & \textbf{+1}        & \textbf{-}          & \textbf{+7}          \\ \hline
\end{tabular}
\hspace*{2 mm}
\begin{tabular}{|l|l|c|c|c|c|c|c|c|c|c|c|c|}
\hline
\multicolumn{2}{|c|}{$\rhd$CCID2014} & 1           & 2           & 3           & 4           & 5           & 6           & 7           & 8  & 9           & 10          & sum         \\ \hline
1              & VIF                 & \textbf{-}  & \textbf{+1} & \textbf{+1} & \textbf{+1} & \textbf{+1} & \textbf{+1} & \textbf{+1} & -1 & \textbf{+1} & 0           & \textbf{+6} \\ \hline
2              & MAD                 & -1          & \textbf{-}  & -1          & -1          & -1          & -1          & -1          & -1 & -1          & -1          & -9          \\ \hline
3              & SR\_SIM             & -1          & \textbf{+1} & \textbf{-}  & -1          & -1          & \textbf{+1} & -1          & -1 & \textbf{+1} & -1          & -3          \\ \hline
4              & FSIM$_c$            & -1          & \textbf{+1} & \textbf{+1} & \textbf{-}  & -1          & \textbf{+1} & 0           & -1 & \textbf{+1} & -1          & 0           \\ \hline
5              & GMSD                & -1          & \textbf{+1} & \textbf{+1} & \textbf{+1} & \textbf{-}  & \textbf{+1} & \textbf{+1} & -1 & \textbf{+1} & -1          & +3          \\ \hline
6              & SFF                 & -1          & \textbf{+1} & -1          & -1          & -1          & \textbf{-}  & -1          & -1 & 0           & -1          & -6          \\ \hline
7              & VSI                 & -1          & \textbf{+1} & \textbf{+1} & 0           & -1          & \textbf{+1} & \textbf{-}  & -1 & \textbf{+1} & -1          & 0           \\ \hline
8              & ADD-GSIM            & \textbf{+1} & \textbf{+1} & \textbf{+1} & \textbf{+1} & \textbf{+1} & \textbf{+1} & \textbf{+1} & -  & \textbf{+1} & \textbf{+1} & \textbf{+9} \\ \hline
9              & SCQI                & -1          & \textbf{+1} & -1          & -1          & -1          & 0           & -1          & -1 & -           & -1          & -6          \\ \hline
10             & MDSI                & 0           & \textbf{+1} & \textbf{+1} & \textbf{+1} & \textbf{+1} & \textbf{+1} & \textbf{+1} & -1 & \textbf{+1} & -           & \textbf{+6} \\ \hline
\end{tabular}
\newline
\vspace*{2 mm}
\newline
\hspace*{-2 mm}
\begin{tabular}{|l|l|c|c|c|c|c|c|c|c|c|c|c|}
\hline
\multicolumn{2}{|c|}{$\rhd$ESPL} & 1           & 2           & 3           & 4           & 5           & 6           & 7           & 8           & 9           & 10 & sum         \\ \hline
1            & VIF               & \textbf{-}  & -1          & -1          & -1          & -1          & -1          & -1          & -1          & -1          & -1 & -9          \\ \hline
2            & MAD               & \textbf{+1} & \textbf{-}  & -1          & -1          & \textbf{+1} & \textbf{+1} & -1          & \textbf{+1} & \textbf{+1} & -1 & +1          \\ \hline
3            & SR\_SIM           & \textbf{+1} & \textbf{+1} & \textbf{-}  & 0           & \textbf{+1} & \textbf{+1} & \textbf{+1} & \textbf{+1} & \textbf{+1} & -1 & \textbf{+6} \\ \hline
4            & FSIM$_c$          & \textbf{+1} & \textbf{+1} & 0           & \textbf{-}  & \textbf{+1} & \textbf{+1} & \textbf{+1} & \textbf{+1} & \textbf{+1} & -1 & \textbf{+6} \\ \hline
5            & GMSD              & +1          & -1          & -1          & -1          & \textbf{-}  & \textbf{+1} & -1          & \textbf{+1} & -1          & -1 & -3          \\ \hline
6            & SFF               & \textbf{+1} & -1          & -1          & -1          & -1          & \textbf{-}  & -1          & \textbf{+1} & -1          & -1 & -5          \\ \hline
7            & VSI               & \textbf{+1} & \textbf{+1} & -1          & -1          & \textbf{+1} & \textbf{+1} & \textbf{-}  & \textbf{+1} & \textbf{+1} & -1 & +3          \\ \hline
8            & ADD-GSIM          & \textbf{+1} & -1          & -1          & -1          & -1          & -1          & -1          & -           & -1          & -1 & -7 \\ \hline
9            & SCQI              & \textbf{+1} & -1          & -1          & -1          & \textbf{+1} & \textbf{+1} & -1          & \textbf{+1} & -           & -1 & -1          \\ \hline
10           & MDSI              & \textbf{+1} & \textbf{+1} & \textbf{+1} & \textbf{+1} & \textbf{+1} & \textbf{+1} & \textbf{+1} & \textbf{+1} & \textbf{+1} & -  & \textbf{+9} \\ \hline
\end{tabular}
\hspace*{2 mm}
\label{my-label}
\begin{tabular}{|l|l|c|c|c|c|c|c|c|c|c|c|c|}
\hline
\multicolumn{2}{|c|}{$\rhd$DRIQ} & 1           & 2           & 3           & 4           & 5           & 6           & 7           & 8           & 9           & 10         & sum         \\ \hline
1            & VIF               & \textbf{-}  & \textbf{+1}           & \textbf{+1} & \textbf{+1} & \textbf{+1} & \textbf{+1} & \textbf{+1} & \textbf{+1} & 0           & -1         & \textbf{+5} \\ \hline
2            & MAD               & -1          & \textbf{-}  & -1          & -1          & -1          & -1          & -1          & -1          & -1          & -1         & -9          \\ \hline
3            & SR\_SIM           & -1          & \textbf{+1} & \textbf{-}  & 0           & 0           & -1          & -1          & -1          & -1          & -1         & -5          \\ \hline
4            & FSIM$_c$          & -1          & \textbf{+1} & 0           & \textbf{-}  & 0           & -1          & -1          & -1          & -1          & -1         & -5          \\ \hline
5            & GMSD              & -1          & \textbf{+1} & 0           & 0           & \textbf{-}  & -1          & -1          & -1          & -1          & -1         & -5          \\ \hline
6            & SFF               & -1          & \textbf{+1} & \textbf{+1} & \textbf{+1} & \textbf{+1} & \textbf{-}  & 0           & \textbf{+1} & -1          & -1         & +2          \\ \hline
7            & VSI               & -1          & \textbf{+1} & \textbf{+1} & \textbf{+1} & \textbf{+1} & 0           & \textbf{-}  & \textbf{+1} & -1          & -1         & +2          \\ \hline
8            & ADD-GSIM          & -1          & \textbf{+1} & \textbf{+1} & \textbf{+1} & \textbf{+1} & -1          & -1          & \textbf{-}  & -1          & -1         & -1          \\ \hline
9            & SCQI              & 0           & \textbf{+1} & \textbf{+1} & \textbf{+1} & \textbf{+1} & \textbf{+1} & \textbf{+1} & \textbf{+1} & \textbf{-}  & -1         & \textbf{+6} \\ \hline
10           & MDSI              & \textbf{+1} & \textbf{+1} & \textbf{+1} & \textbf{+1} & \textbf{+1} & \textbf{+1} & \textbf{+1} & \textbf{+1} & \textbf{+1} & \textbf{-} & \textbf{+9} \\ \hline
\end{tabular} }
\end{table*}

\begin{figure}[htb]
\scriptsize
\begin{minipage}[b]{0.49\linewidth}
  \centering
  \centerline{\includegraphics[height=3.5cm]{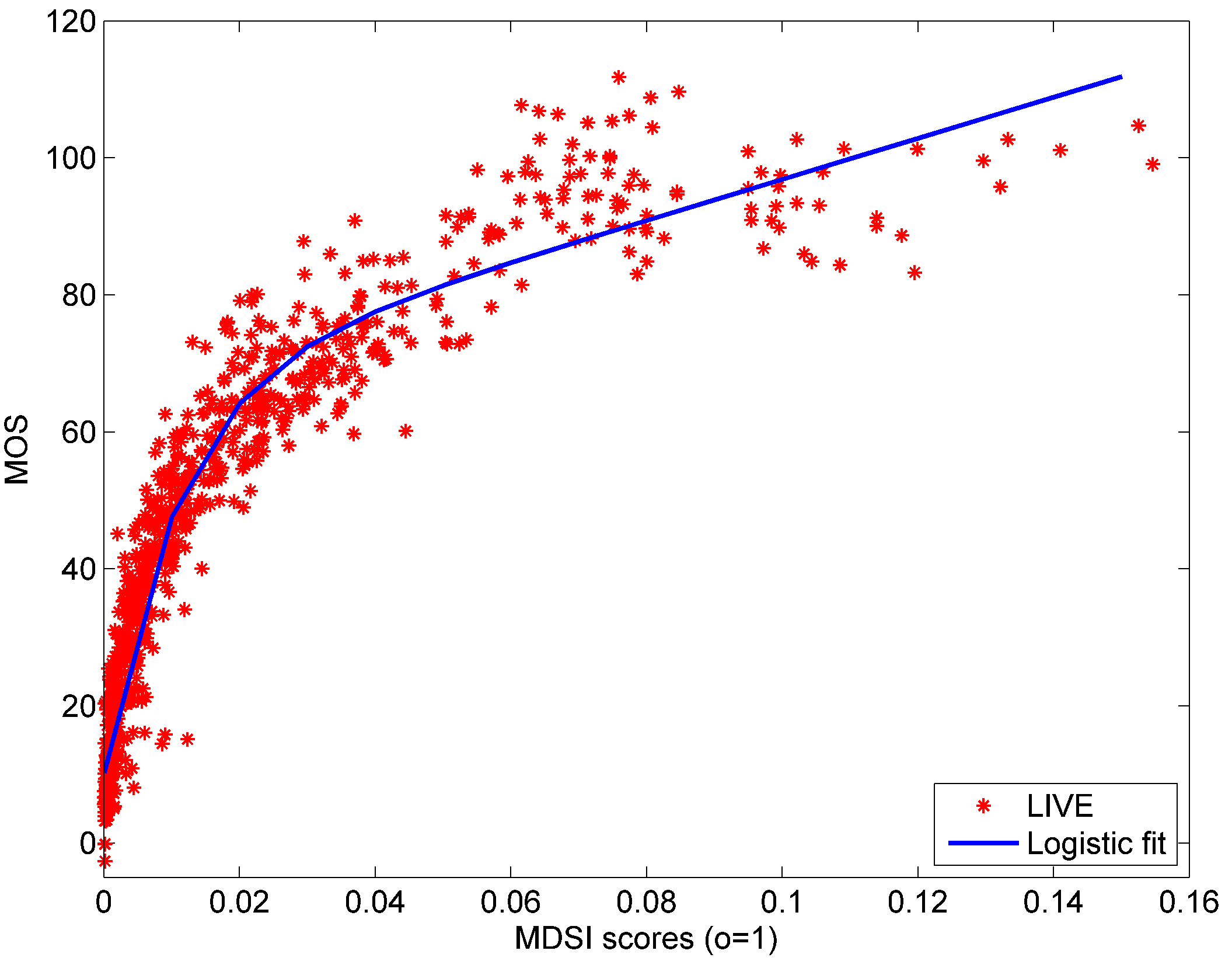}}
\centerline{LPCC = 0.8324}
\centerline{PCC = 0.9626}
\end{minipage}
\begin{minipage}[b]{0.49\linewidth}
  \centering
  \centerline{\includegraphics[height=3.5cm]{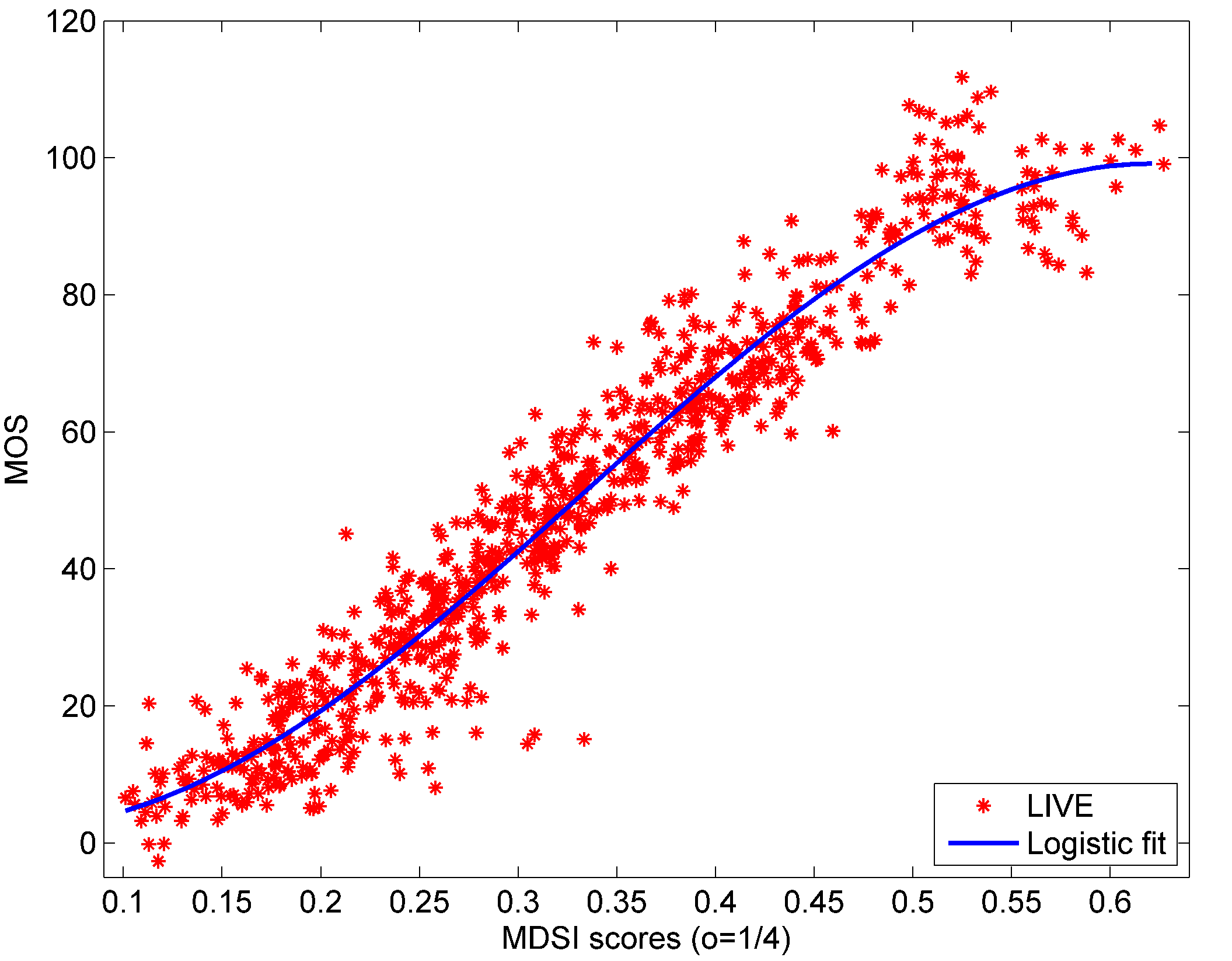}}
\centerline{LPCC = 0.9618}
\centerline{PCC = 0.9659}
\end{minipage}
\caption{Scatter plots of quality scores against the subjective MOS on the LIVE dataset for the proposed model MDSI with and without using the power pooling. Comparison of LPCC and PCC values indicate that MDSI becomes more linear with respect to MOS (the right plot) by using the power pooling.}
\label{scatter}
\end{figure}

For the purpose of visualizing quality scores of the proposed index, the scatter plots of the proposed IQA model MDSI with and without using power pooling are shown in Fig. \ref{scatter}. The logistic function suggested in \cite{statistical2006} was used to fit a curve on each plot:

\begin{equation}
  \ f(x) = \beta_1\Big(\frac{1}{2}-\frac{1}{1-e^{\beta_2(x-\beta_3)}}\Big)+\beta_4x+\beta_5
  \label{equ:REG}
\end{equation}                    
where $\beta_1$, $\beta_2$, $\beta_3$, $\beta_4$ and $\beta_5$ are fitting parameters computed by minimizing the mean square error between quality predictions $x$ and subjective scores MOS. It should be noted that reported PCC and RMSE values in this paper are computed after mapping quality scores to MOS based on above function.

The reported results in Table \ref{results1} show the difference between different IQA models. As suggested in \cite{VQEG2003, statistical2006}, we use F-test to decide whether a metric is statistically superior to another index. The F-test is based on the residuals between the quality scores given by an IQA model after applying nonlinear mapping of equation (\ref{equ:REG}), and the mean subjective scores MOS. The ratio of variance between residual errors of an IQA model to another model at 95\% significance level is used by F-test. The result of the test is equal to 1 if we can reject the null hypothesis and 0 otherwise. The results of F-test on eight datasets are listed in Table \ref{Ftest}. In this Table, +1/-1 indicate that corresponding index is statistically superior/inferior to the other index being compared to. If the difference between two indices is not significant, the result is shown by 0. We note that type I error might be occurred, specially when quality scores of IQA models are not Gaussian. However, even existence of possible errors is very unlikely to result in another conclusion about the superiority of the proposed index because there is a considerable gap between the proposed index and the other metrics as discussed in the following.

From the results of Table \ref{Ftest}, we can see that MDSI is significantly better than the other indices on TID2008, TID2013, 	ESPL, and DRIQ datasets. Therefore, its sum value in the last column is +9 for these four datasets. SCQI is statistically superior to the other indices on the TID2013 dataset except for MDSI. On the LIVE dataset, indices MAD, MDSI, and ADD-GSIM are significantly better than the other indices. On the CSIQ dataset, only SFF performs significantly better than MDSI. On the CCID2014 dataset, ADD-GSIM is significantly better than the other indices, while the statistically indistinguishable indices VIF and MDSI show promising results. Considering all eight datasets used in this experiment, with a minimum sum value of +6, the proposed index MDSI performs very well in comparison with the other indices. We can simply add the eight cumulative sum values of each metric for the eight datasets to have an overall comparison based on the statistical significance test. This score indicates how many times a metric is statistically superior to the other metrics. The results show that MDSI is the best performing index by a score of +62 (out of maximum +72), followed by ADD-GSIM (+25), VSI (+1), SCQI (-2), FSIM$_c$ (-4), GMSD (-5), SFF (-6), SR\_SIM (-19), MAD (-24), and VIF (-26). The results based on the statistical significance test verify that unlike other IQA models, the proposed metric MDSI is among the best performing indices on different datasets.

\subsection{Performance comparison on individual distortions}
\label{individual}

A good IQA model should perform not only accurate quality predictions for a whole dataset; it should provide good judgments over individual distortion types. We list in Table \ref{results2} the average SRC, and PCC values of thirteen IQA models for 61 sets of distortions available in the six datasets of TID2008, CSIQ, LIVE, TID2013, VCL@FER, and ESPL. The minimum value for each evaluation metric and standard deviation of these 61 values are also listed. These two evaluations indicate to the \textit{reliability} of an IQA model. An IQA model should provide good prediction accuracy for all of the distortion types. If a metric fails at assessing one or more types of distortions, that index can not be reliable.   

The proposed index MDSI, has the best SRC, and PCC average on distortion types. MDSI, SCQI and FSIM$_c$ in the worst case perform better than the other IQA models, as can be seen in the \textit{min} column for each evaluation metric. This shows the reliability of the proposed index. The negative \textit{min} values and close to zero \textit{min} values in Table \ref{results2} indicate the unreliability of related models when dealing with some distortion types. The standard deviation of 61 values for each evaluation metric is another reliability factor. According to Table \ref{results2}, MDSI, SCQI and FSIM$_c$ have the lowest variation. Therefore, we can conclude that indices MDSI, SCQI and FSIM$_c$ are more reliable than the other indices.

\begin{table}[htb]
\caption{Overall performance comparison of the proposed IQA model MDSI and twelve popular/competing indices on individual distortion types of six datasets (TID2008, CSIQ, LIVE, TID2013, VCL@FER, and ESPL). The six datasets contain 61 distortion set, therefore results on distortion types are reported based on average of 61 correlation values. Top three IQA models are highlighted.}
\scriptsize
\centering
\begin{tabular}{c|ccc|ccc}
\hline
\multirow{2}{*}{IQA model} & \multicolumn{3}{c|}{SRC (Distortions)}              & \multicolumn{3}{c}{PCC (Distortions)}               \\ \cline{2-7} 
                           & avg             & min             & std             & avg             & min             & std             \\ \hline
MSSSIM                     & 0.8343          & -0.4099         & 0.1989          & 0.8560          & -0.4448         & 0.1944          \\
VIF                        & 0.8537          & -0.3099         & 0.1811          & 0.8760          & -0.3443         & 0.1812          \\
MAD                        & 0.8111          & -0.0575         & 0.2315          & 0.8296          & 0.0417          & 0.2108          \\
IWSSIM                     & 0.8329          & -0.4196         & 0.2019          & 0.8568          & -0.4503         & 0.1962          \\
SR\_SIM                    & 0.8609          & -0.2053         & 0.1806          & 0.8785          & -0.3162         & 0.1839          \\
FSIM$_c$                      & 0.8775          & \textbf{0.4679} & \textbf{0.1041} & 0.8967          & \textbf{0.5488} & \textbf{0.0880} \\
GMSD                       & 0.8542          & -0.2948         & 0.1954          & 0.8785          & -0.3625         & 0.1851          \\
SFF                        & 0.8538          & 0.1786          & 0.1472          & 0.8721          & 0.0786          & 0.1441          \\
VSI                        & \textbf{0.8779} & 0.1713          & 0.1360          & \textbf{0.8969} & 0.4875          & 0.1044          \\
DSCSI                      & 0.8722          & 0.3534          & 0.1242          & 0.8908          & 0.5166          & 0.1093          \\
ADD-GSIM                   & 0.8650          & -0.2053         & 0.1686          & 0.8799          & -0.2190         & 0.1691          \\
SCQI                       & \textbf{0.8826} & \textbf{0.4479} & \textbf{0.1057} & \textbf{0.9010} & \textbf{0.6493} & \textbf{0.0841} \\
MDSI                       & \textbf{0.8903} & \textbf{0.4378} & \textbf{0.1030} & \textbf{0.9095} & \textbf{0.6899} & \textbf{0.0805} \\ \hline
\end{tabular}
\label{results2}
\end{table}

\begin{table}[htb]
\centering
\scriptsize
\caption{Performance of the proposed index MDSI with different pooling strategies and values of parameter $q$.}
\begin{tabular}{c|ccc|ccc}
\hline
\multirow{2}{*}{Pooling} & \multicolumn{3}{c|}{Weighted avg. SRC (8 datasets)} & \multicolumn{3}{c}{Avg. SRC (61 Distortions)} \\ \cline{2-7} 
                         & Mean      & MAD                & SD                 & Mean                & MAD        & SD         \\ \hline
q = 1/4                  & 0.8864    & \textbf{0.9066}    & 0.8776             & \textbf{0.8919}     & 0.8903     & 0.8828     \\
q = 1/2                  & 0.8833    & \textbf{0.9067}    & 0.8820             & \textbf{0.8912}     & 0.8898     & 0.8826     \\
q = 1                    & 0.8730    & \textbf{0.9041}    & 0.8899             & \textbf{0.8899}     & 0.8890     & 0.8820     \\
q = 2                    & 0.8519    & 0.8928             & \textbf{0.8972}    & \textbf{0.8888}     & 0.8866     & 0.8820     \\
q = 4                    & 0.8301    & 0.8766             & \textbf{0.8922}    & \textbf{0.8869}     & 0.8780     & 0.8753     \\ \hline
\end{tabular}
\label{meanmadstd}
\end{table}

\subsection{Parameters of deviation pooling ($\rho$, $q$, $o$)}
\label{vs1}

Considering the formulation of deviation pooling in equation (\ref{equ:DP}), we used the mean absolute deviation (MAD), e.g. $\rho=1$, for the proposed metric. Standard deviation (SD), e.g. $\rho=2$, is another option that can be used for deviation pooling. In addition, the Minkowski power ($q$) of the deviation pooling can have significant impact on the proposed index. In Table \ref{meanmadstd}, the SRC performance of the proposed index is analyzed for different values of $q$ and $\rho=\{1, 2\}$. Mean pooling is also used in this experiment. The results show that MAD pooling with $q<=1$ is a better choice for the proposed index. Also, the performance of the mean pooling on 61 distortion set confirms our statement that mean pooling has a good performance for inter-class quality prediction.

The impact of the proposed power pooling of the deviation pooling on the proposed metric was shown in Fig. \ref{scatter}. Power pooling can be also used to increase linearity of other indices as well. For example, LPCC and PCC values of VSI \cite{VSI} for TID2013 dataset, by setting $o=18$, can be increased from 0.8373 to 0.8928, and 0.9000 to 0.9011, respectively.

\subsection{Summation vs. Multiplication}
\label{vs2}
    
Two options for combination of the two similarity maps GS/$\widehat{\text{GS}}$ and $\widehat{\text{CS}}$ are summation and multiplication as explained in subsection \ref{Chromaticity}. Deciding whether one approach is superior to another for an index depends on many factors. These factors might be the pooling strategy being used, overall performance, performance on individual distortions, reliability, efficiency, simplicity, etc. In an experiment, the performance of the MDSI using the multiplication approach was examined. Based on the many set of parameters were tested, we found that $\gamma = 0.2$ and $\beta = 0.1$ are good parameters to combine $\widehat{\text{GS}}$ and $\widehat{\text{CS}}$ via the multiplication scheme. The observation was that summation is a better choice for TID2008, TID2013, VCL@FER, and DRIQ datasets, while multiplication is a better choice for ESPL dataset, and that both approaches show almost the same performance on other datasets. Overall, the summation approach provides better performance on individual distortions. This experiment also shown that MDSI is more reliable through summation than multiplication based on the reliability measures introduced in this paper. Based on this experiment, the simplicity of the summation combination approach and its efficiency over multiplication, the former was used along with MDSI. Table \ref{decide} justifies our choice.

\begin{table}[htb]
\centering
\scriptsize
\caption{Different criteria used to choose the combination scheme.}
\begin{tabular}{c|cc}
Property                                         & Summation                      & Multiplication       \\ \hline
Statistically superior over more considered datasets        & \textbf{\checkmark}                                    &   \\
Better dataset-weighted average                  & \checkmark                     & \textbf{}            \\
Better performance on individual distortions     & \checkmark                     & \textbf{}            \\
Reliability                                      & \checkmark                     &                      \\
Simplicity                                       & \checkmark 
&                      \\
Efficiency                                       & \checkmark                     &                      \\ \hline
\end{tabular}
\label{decide}
\end{table}

\subsection{Parameters of model}
\label{parameters}

The proposed IQA model MDSI has four parameters to be set. The four parameters of MDSI are $C_1$, $C_2$, $C_3$ and $\alpha$. To further simplify the MDSI, we set $C_3 = 4C_1 = 10C_2$. Therefore, MDSI has only two parameters to set, e.g. $C_3$ and $\alpha$. For an example, we refer to the SSIM index \cite{SSIM} that also uses such a simplification. Note that gradient similarities and chromaticity similarity have different dynamic ranges, therefore, these parameters should be set such that the relation between these maps also be taken into account. 

\begin{figure}[htb]
\scriptsize
\begin{minipage}[b]{0.99\linewidth}
  \centering
  \centerline{\includegraphics[height=5.2cm]{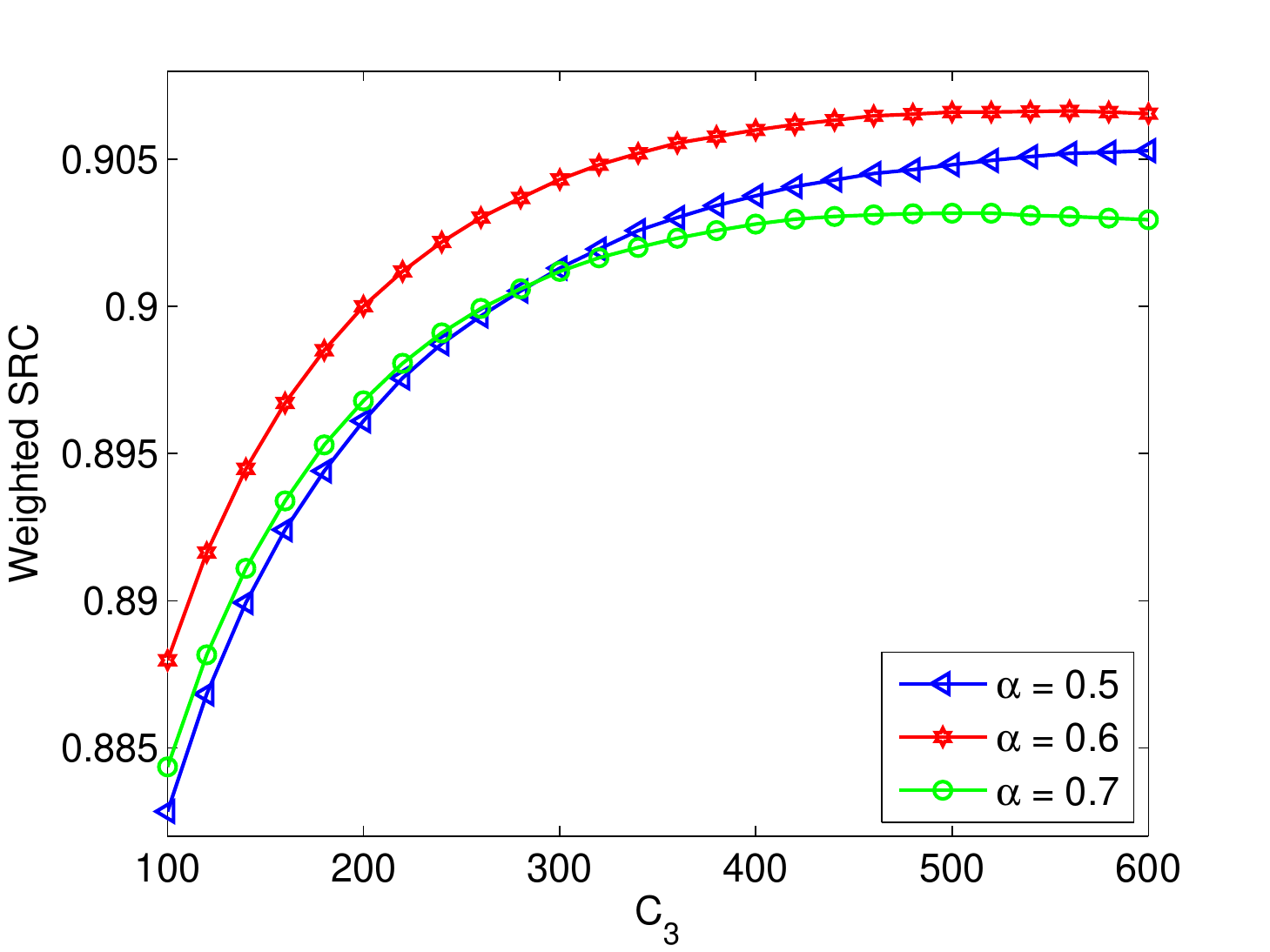}} 
\end{minipage}
\caption{The weighted SRC performance of MDSI for different values of $C_3$ and $\alpha$ on eight datasets (TID2008, CSIQ, LIVE, TID2013, VCL@FER, CCID2014, ESPL, and DRIQ).}
\label{fig:parameters2}
\end{figure}

In Fig. \ref{fig:parameters2}, the impact of these two parameters on the performance of the MDSI is shown. Even though the parameters $C_1$, $C_2$ and $C_3$ are set approximately, it can be seen that MDSI is very robust under different setup of parameters. MDSI has greater weighted average SRC than 0.90 for any $\alpha \in [0.5 ~ 0.7]$ and $C_3 \in [300 ~ 600]$. Note that many other possible setup of parameters are not included in this plot. In the experiments, we set $\alpha = 0.6$, $C_1 = 140$, $C_2 = 55$, and $C_3 = 550$.

\subsection{Effect of chromaticity similarity maps CS and $\widehat{\text{CS}}$}

\begin{figure}[htb]
\scriptsize
\begin{minipage}[b]{.99\linewidth}
  \centering
  \centerline{\includegraphics[height=5.2cm]{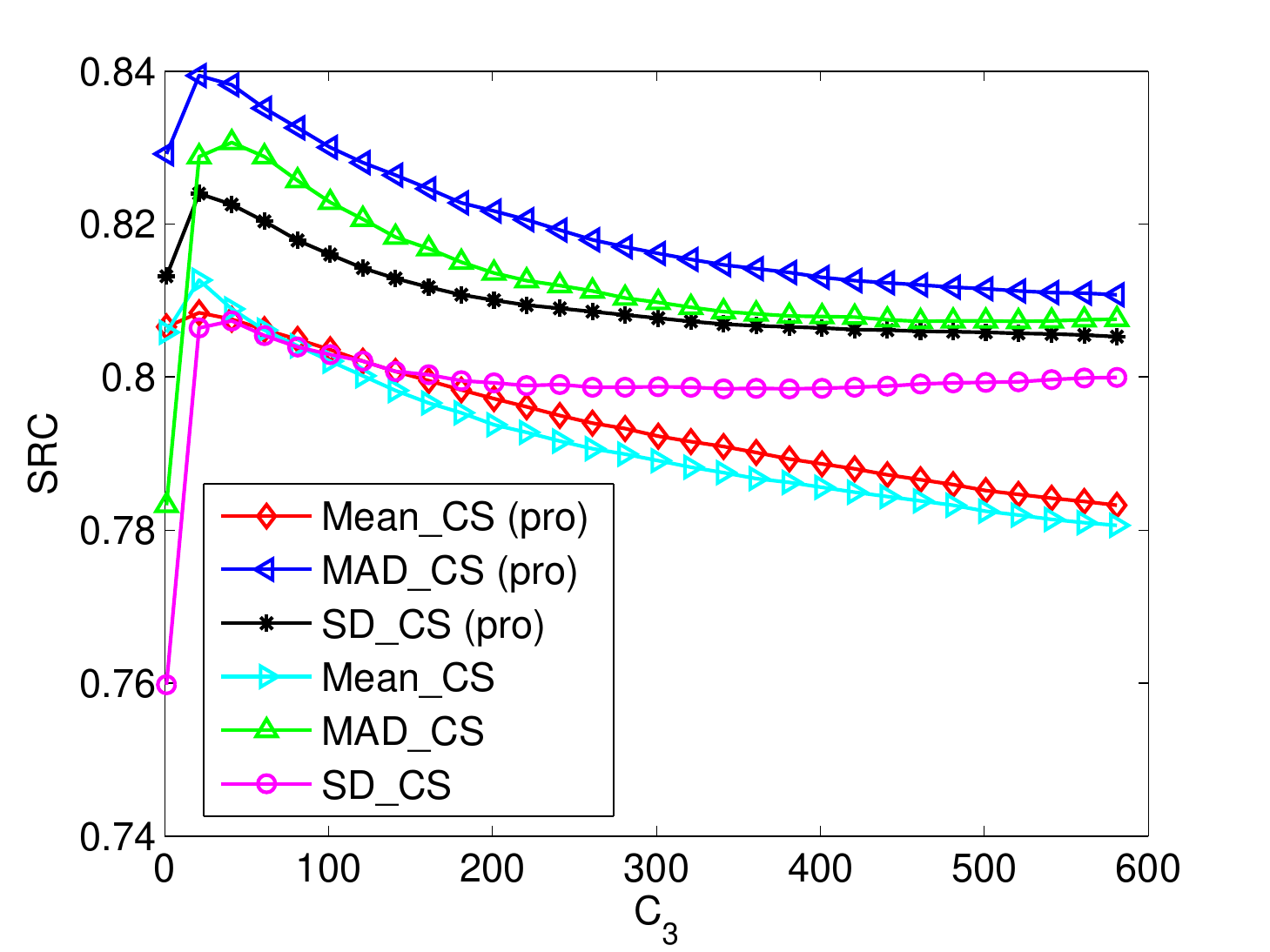}}
\end{minipage}
\caption{The SRC performance of the proposed index MDSI with two chromaticity similarity maps CS and $\widehat{\text{CS}}$ (proposed) for different values of $C_3$ and three pooling strategies on CCID2014 dataset \cite{CCID2014}.}
\label{CSJCS}
\end{figure}

In this section, the impact of using CS \cite{VSI} and proposed $\widehat{\text{CS}}$ on the performance of the proposed index is studied through the following experiment. Contrast distorted images of the CCID2014 dataset \cite{CCID2014} were chosen. The reason of choosing this dataset is to evaluate the ability of measuring color changes by CS and $\widehat{\text{CS}}$. We analyzed the SRC performance of the CS and $\widehat{\text{CS}}$ as a part of the proposed index for wide range of $C_3$ values. Three pooling strategies were used in this experiment, e.g. mean pooling, mean absolute deviation (MAD) pooling and the standard deviation (SD) pooling. Fig. \ref{CSJCS} shows the SRC performance of the proposed index for different scenarios. From the plot in Fig. \ref{CSJCS}, the following conclusions can be drawn. MAD pooling and both CS and $\widehat{\text{CS}}$ are good choices for MDSI. For almost every pooling strategy and parameter of $C_3$, the proposed $\widehat{\text{CS}}$ performs better than CS. This advantage is at the same time that the proposed $\widehat{\text{CS}}$ is more efficient than the existing CS.

\subsection{Implementation and efficiency}
\label{efficiency}

Another very important factor of a good IQA model is its efficiency. The proposed index has a very low complexity. It first applies average filtering of size $M \times M$ on each channel of the $\mathcal{R}$ and $\mathcal{D}$ images, downsample them by a factor of $M$ and convert the results to a luminance and two chromaticity channels \cite{note2016}. The value of $M$ is set to $[min(h, w)/256]$ \cite{Wangsite}, where $h$ and $w$ are image height and width, and $[.]$ is the round operator. Then, the proposed index calculates the gradient magnitudes of luminance channel, the chromaticity similarity map, and apply deviation pooling. All these steps are computationally efficient. Table \ref{time} lists the run times of fifteen IQA models when applied on images of size 384$\times$512 and 1080$\times$1920. The experiments were performed on a Corei7 3.40GHz CPU with 16 GB of RAM. The IQA models were implemented in MATLAB 2013b running on Windows 7. It can be seen that MDSI is among top five fastest indices. The proposed index is less than 2 times slower than the competing GMSD index. The reason for this is that GMSD only uses the luminance feature. Compared to the other competing indices, SCQI, VSI, ADD-GSIM, SFF, and FSIM$_c$, the proposed index MDSI is about 3 to 6 times, 3 to 9 times, 4 to 5 times, 4 to 5 times, and 4 to 11 times faster, respectively. Another observation from the Table \ref{time} is that the ranking of indices might not be the same when they are tested on images of different size. For example, SSIM performs slower than the proposed index on smaller images, but faster on larger images.

\begin{table}[htb]
\centering
\scriptsize
\caption{Run time comparison of IQA models in terms of milliseconds}
\begin{tabular}{lcc}
\hline
IQA model                    & 384$\times$512 & 1080$\times$1920 \\ \hline
{PSNR}    & 5.69 & 37.85                       \\
{GMSD} \cite{GMSD}    & 8.90 & 78.22                      \\
$\rhd$ {MDSI}    & 12.21 & 152.85                       \\
{SSIM} \cite{SSIM}    & 14.97 & 80.23                       \\
{SR\_SIM} \cite{SRSIM} & 17.02 & 100.06                       \\
{MSSSIM} \cite{MSSSIM}  & 52.16 & 413.70                       \\
{ADD-GSIM} \cite{ADD-GSIM}  & 59.58 & 566.99                    \\
{SFF} \cite{SFF}    & 64.22 & 588.57                       \\
{SCQI} \cite{SCQI}    & 71.68 & 524.01                       \\
{VSI} \cite{VSI}    & 106.87 & 492.85                      \\
{FSIM$_c$} \cite{FSIM}   & 145.02 & 590.84                     \\
{IWSSIM} \cite{IWSSIM} & 244.00 & 2538.43                      \\
{DSCSI} \cite{DSCSI}    & 423.73 & 4599.83                       \\
{VIF} \cite{VIF}    & 635.22 & 6348.67                      \\
{MAD} \cite{MAD}    & 847.54 & 8452.50                     \\
\hline   
\end{tabular}
\label{time}
\end{table}


\section{Conclusion}
\label{conclusion}

We propose an effective, efficient, and reliable full reference IQA model based on the new gradient and chromaticity similarities. The gradient similarity was used to measure local structural distortions. In a complementary way, a chromaticity similarity was proposed to measure color distortions. The proposed metric, called MDSI, use a novel deviation pooling to compute the quality score from the two similarity maps. Extensive experimental results on natural and synthetic benchmark datasets prove that the proposed index is effective and reliable, has low complexity, and is fast enough to be used in real-time FR-IQA applications.

\section*{Acknowledgments}
The authors thank the NSERC of Canada for their financial support under Grants RGPDD 451272-13 and RGPIN 138344-14.



\bibliographystyle{IEEEtran}
\bibliography{egbib2}   

\begin{thebibliography}{10}
\providecommand{\url}[1]{#1}
\csname url@samestyle\endcsname
\providecommand{\newblock}{\relax}
\providecommand{\bibinfo}[2]{#2}
\providecommand{\BIBentrySTDinterwordspacing}{\spaceskip=0pt\relax}
\providecommand{\BIBentryALTinterwordstretchfactor}{4}
\providecommand{\BIBentryALTinterwordspacing}{\spaceskip=\fontdimen2\font plus
\BIBentryALTinterwordstretchfactor\fontdimen3\font minus
  \fontdimen4\font\relax}
\providecommand{\BIBforeignlanguage}[2]{{%
\expandafter\ifx\csname l@#1\endcsname\relax
\typeout{** WARNING: IEEEtran.bst: No hyphenation pattern has been}%
\typeout{** loaded for the language `#1'. Using the pattern for}%
\typeout{** the default language instead.}%
\else
\language=\csname l@#1\endcsname
\fi
#2}}
\providecommand{\BIBdecl}{\relax}
\BIBdecl

\bibitem{applications2011}
Z.~Wang, ``Applications of objective image quality assessment methods
  [applications corner],'' \emph{IEEE Signal Processing Magazine}, vol.~28,
  no.~6, pp. 137--142, Nov 2011.

\bibitem{SSIM}
Z.~Wang, A.~Bovik, H.~Sheikh, and E.~Simoncelli, ``Image quality assessment:
  from error visibility to structural similarity,'' \emph{IEEE Transactions on
  Image Processing}, vol.~13, no.~4, pp. 600--612, April 2004.

\bibitem{TMQI}
H.~Yeganeh and Z.~Wang, ``Objective quality assessment of tone-mapped images,''
  \emph{IEEE Transactions on Image Processing}, vol.~22, no.~2, pp. 657--667,
  Feb 2013.

\bibitem{FSITM}
H.~Ziaei~Nafchi, A.~Shahkolaei, R.~Farrahi~Moghaddam, and M.~Cheriet,
  ``{{FSITM: A Feature Similarity Index For Tone-Mapped Images}},'' \emph{IEEE
  Signal Processing Letters}, vol.~22, no.~8, pp. 1026--1029, Aug 2015.

\bibitem{spm2009}
Z.~Wang and A.~Bovik, ``{{Mean squared error: Love it or leave it? A new look
  at Signal Fidelity Measures}},'' \emph{IEEE Signal Processing Magazine},
  vol.~26, no.~1, pp. 98--117, Jan 2009.

\bibitem{NQM}
N.~Damera-Venkata, T.~Kite, W.~Geisler, B.~Evans, and A.~Bovik, ``Image quality
  assessment based on a degradation model,'' \emph{IEEE Transactions on Image
  Processing}, vol.~9, no.~4, pp. 636--650, Apr 2000.

\bibitem{MSSSIM}
Z.~Wang, E.~Simoncelli, and A.~Bovik, ``Multiscale structural similarity for
  image quality assessment,'' in \emph{Thirty-Seventh Asilomar Conference on
  Signals, Systems and Computers}, vol.~2, Nov 2003, pp. 1398--1402.

\bibitem{IFC}
H.~Sheikh, A.~Bovik, and G.~de~Veciana, ``An information fidelity criterion for
  image quality assessment using natural scene statistics,'' \emph{IEEE
  Transactions on Image Processing}, vol.~14, no.~12, pp. 2117--2128, Dec 2005.

\bibitem{VIF}
H.~Sheikh and A.~Bovik, ``Image information and visual quality,'' \emph{IEEE
  Transactions on Image Processing}, vol.~15, no.~2, pp. 430--444, Feb 2006.

\bibitem{VSNR}
D.~Chandler and S.~Hemami, ``{{VSNR: A Wavelet-Based Visual Signal-to-Noise
  Ratio for Natural Images}},'' \emph{IEEE Transactions on Image Processing},
  vol.~16, no.~9, pp. 2284--2298, Sept 2007.

\bibitem{RFSIM}
L.~Zhang, D.~Zhang, and X.~Mou, ``{{RFSIM: A feature based image quality
  assessment metric using Riesz transforms}},'' in \emph{17th IEEE
  International Conference on Image Processing (ICIP)}, Sept 2010, pp.
  321--324.

\bibitem{MAD}
E.~C. Larson and D.~M. Chandler, ``Most apparent distortion: full-reference
  image quality assessment and the role of strategy,'' \emph{Journal of
  Electronic Imaging}, vol.~19, no.~1, p. 011006, 2010.

\bibitem{SVR2010}
M.~Narwaria and W.~Lin, ``Objective image quality assessment based on support
  vector regression,'' \emph{IEEE Transactions on Neural Networks}, vol.~21,
  no.~3, pp. 515--519, March 2010.

\bibitem{CPSSIM}
C.~Li and A.~C. Bovik, ``Content-partitioned structural similarity index for
  image quality assessment,'' \emph{Signal Processing: Image Communication},
  vol.~25, no.~7, pp. 517--526, 2010, special Issue on Image and Video Quality
  Assessment.

\bibitem{IWSSIM}
Z.~Wang and Q.~Li, ``Information content weighting for perceptual image quality
  assessment,'' \emph{IEEE Transactions on Image Processing}, vol.~20, no.~5,
  pp. 1185--1198, May 2011.

\bibitem{FSIM}
L.~Zhang, D.~Zhang, X.~Mou, and D.~Zhang, ``{{FSIM: A Feature Similarity Index
  for Image Quality Assessment}},'' \emph{IEEE Transactions on Image
  Processing}, vol.~20, no.~8, pp. 2378--2386, Aug 2011.

\bibitem{GS}
A.~Liu, W.~Lin, and M.~Narwaria, ``Image quality assessment based on gradient
  similarity,'' \emph{IEEE Transactions on Image Processing}, vol.~21, no.~4,
  pp. 1500--1512, April 2012.

\bibitem{SRSIM}
L.~Zhang and H.~Li, ``{{SR-SIM: A fast and high performance IQA index based on
  spectral residual}},'' in \emph{19th IEEE International Conference on Image
  Processing (ICIP)}, Sept 2012, pp. 1473--1476.

\bibitem{GMSD}
W.~Xue, L.~Zhang, X.~Mou, and A.~Bovik, ``{{Gradient Magnitude Similarity
  Deviation: A Highly Efficient Perceptual Image Quality Index}},'' \emph{IEEE
  Transactions on Image Processing}, vol.~23, no.~2, pp. 684--695, Feb 2014.

\bibitem{SFF}
H.-W. Chang, H.~Yang, Y.~Gan, and M.-H. Wang, ``{{Sparse Feature Fidelity for
  Perceptual Image Quality Assessment}},'' \emph{IEEE Transactions on Image
  Processing}, vol.~22, no.~10, pp. 4007--4018, Oct 2013.

\bibitem{VSI}
L.~Zhang, Y.~Shen, and H.~Li, ``{{VSI: A Visual Saliency-Induced Index for
  Perceptual Image Quality Assessment}},'' \emph{IEEE Transactions on Image
  Processing}, vol.~23, no.~10, pp. 4270--4281, Oct 2014.

\bibitem{ADD-GSIM}
K.~Gu, S.~Wang, G.~Zhai, W.~Lin, X.~Yang, and W.~Zhang, ``Analysis of
  distortion distribution for pooling in image quality prediction,'' \emph{IEEE
  Transactions on Broadcasting}, vol.~62, no.~2, pp. 446--456, June 2016.

\bibitem{SCQI}
S.~H. Bae and M.~Kim, ``A novel image quality assessment with globally and
  locally consilient visual quality perception,'' \emph{IEEE Transactions on
  Image Processing}, vol.~25, no.~5, pp. 2392--2406, 2016.

\bibitem{Fusion2016}
M.~Oszust, ``Decision fusion for image quality assessment using an optimization
  approach,'' \emph{IEEE Signal Processing Letters}, vol.~23, no.~1, pp.
  65--69, Jan 2016.

\bibitem{Metrics2011}
W.~Lin and C.-C.~J. Kuo, ``Perceptual visual quality metrics: A survey,''
  \emph{Journal of Visual Communication and Image Representation}, vol.~22,
  no.~4, pp. 297 -- 312, 2011.

\bibitem{SVD2006}
A.~Shnayderman, A.~Gusev, and A.~Eskicioglu, ``{{An SVD-based grayscale image
  quality measure for local and global assessment}},'' \emph{IEEE Transactions
  on Image Processing}, vol.~15, no.~2, pp. 422--429, Feb 2006.

\bibitem{wavelet2011}
S.~Li, F.~Zhang, L.~Ma, and K.~N. Ngan, ``Image quality assessment by
  separately evaluating detail losses and additive impairments,'' \emph{IEEE
  Transactions on Multimedia}, vol.~13, no.~5, pp. 935--949, Oct 2011.

\bibitem{CWSSIM}
M.~Sampat, Z.~Wang, S.~Gupta, A.~Bovik, and M.~Markey, ``Complex wavelet
  structural similarity: A new image similarity index,'' \emph{IEEE
  Transactions on Image Processing}, vol.~18, no.~11, pp. 2385--2401, Nov 2009.

\bibitem{spatial2006}
Z.~Wang and X.~Shang, ``Spatial pooling strategies for perceptual image quality
  assessment,'' in \emph{IEEE International Conference on Image Processing},
  Oct 2006, pp. 2945--2948.

\bibitem{pooling2009}
A.~Moorthy and A.~Bovik, ``Visual importance pooling for image quality
  assessment,'' \emph{IEEE Journal of Selected Topics in Signal Processing},
  vol.~3, no.~2, pp. 193--201, April 2009.

\bibitem{percentile2009}
A.~K. Moorthy and A.~C. Bovik, ``Perceptually significant spatial pooling
  techniques for image quality assessment,'' \emph{Proc. SPIE}, vol. 7240, pp.
  724\,012--724\,012--11, 2009.

\bibitem{GSSIM}
G.-H. Chen, C.-L. Yang, and S.-L. Xie, ``Gradient-based structural similarity
  for image quality assessment,'' in \emph{IEEE International Conference on
  Image Processing}, Oct 2006, pp. 2929--2932.

\bibitem{gradient2010}
D.-O. Kim, H.-S. Han, and R.-H. Park, ``Gradient information-based image
  quality metric,'' \emph{IEEE Transactions on Consumer Electronics}, vol.~56,
  no.~2, pp. 930--936, May 2010.

\bibitem{invariance2001}
J.-M. Geusebroek, R.~Van~den Boomgaard, A.~Smeulders, and H.~Geerts, ``Color
  invariance,'' \emph{IEEE Transactions on Pattern Analysis and Machine
  Intelligence}, vol.~23, no.~12, pp. 1338--1350, Dec 2001.

\bibitem{sobel}
G.~F. I.~Sobel, ``A 3x3 isotropic gradient operator for image processing,''
  1968, presented at a talk at the Stanford Artificial Project.

\bibitem{ITUT2012}
I.-T.~P. 1401, ``Methods, metrics and procedures for statistical evaluation,
  qualification and comparison of objective quality prediction models,''
  \url{https://www.itu.int/rec/dologin_pub.asp?lang=e&id=T-REC-P.1401-201207-I!!PDF-E&type=items},
  July 2012.

\bibitem{LIVEweb}
H.~Sheikh, Z.~Wang, L.~Cormack, and A.~Bovik, ``Live image quality assessment
  database release 2,'' Online:
  \url{http://live.ece.utexas.edu/research/quality}.

\bibitem{TID2008}
N.~Ponomarenko, V.~Lukin, A.~Zelensky, K.~Egiazarian, M.~Carli, and
  F.~Battisti, ``{{TID2008 - A Database for Evaluation of Full-Reference Visual
  Quality Assessment Metrics}},'' \emph{Advances of Modern Radioelectronics},
  vol.~10, pp. 30--45, 2009.

\bibitem{TID2013}
N.~Ponomarenko, O.~Ieremeiev, V.~Lukin, K.~Egiazarian, L.~Jin, J.~Astola,
  B.~Vozel, K.~Chehdi, M.~Carli, F.~Battisti, and C.-C. Kuo, ``{{Color image
  database TID2013: Peculiarities and preliminary results}},'' in \emph{4th
  European Workshop on Visual Information Processing (EUVIP)}, June 2013, pp.
  106--111.

\bibitem{VCL}
N.~B. H. H. M. L. E. D. S.~G. A.~Zaric, N.~Tatalovic, ``{{VCL@FER Image Quality
  Assessment Database}},'' \emph{AUTOMATIKA}, vol.~53, no.~4, pp. 344--354,
  2012.

\bibitem{CCID2014}
K.~Gu, G.~Zhai, W.~Lin, and M.~Liu, ``The analysis of image contrast: From
  quality assessment to automatic enhancement,'' \emph{IEEE Transactions on
  Cybernetics}, vol.~46, no.~1, pp. 284--297, 2015.

\bibitem{ESPL}
D.~Kundu and B.~L. Evans, ``Full-reference visual quality assessment for
  synthetic images: A subjective study,'' in \emph{Proc. IEEE Int. Conf. on
  Image Processing}, 2015, pp. 2374--2378.

\bibitem{DRIQ}
C.~Vu, T.~Phan, P.~Singh, , and D.~M. Chandler, ``{{Digitally Retouched Image
  Quality (DRIQ) Database}},'' Online: \url{http://vision.okstate.edu/driq/},
  2012.

\bibitem{DSCSI}
D.~Lee and K.~N. Plataniotis, ``Towards a full-reference quality assessment for
  color images using directional statistics,'' \emph{IEEE Transactions on Image
  Processing}, vol.~24, no.~11, pp. 3950--3965, Nov 2015.

\bibitem{statistical2006}
H.~Sheikh, M.~Sabir, and A.~Bovik, ``A statistical evaluation of recent full
  reference image quality assessment algorithms,'' \emph{IEEE Transactions on
  Image Processing}, vol.~15, no.~11, pp. 3440--3451, Nov 2006.

\bibitem{VQEG2003}
``Final report from the video quality experts group on the validation of
  objective models of video quality assessment,''
  \url{ftp://vqeg.its.bldrdoc.gov/Documents/Meetings/Hillsboro_VQEG_Mar_03/VQEGIIDraftReportv2a.pdf},
  2003.

\bibitem{note2016}
\BIBentryALTinterwordspacing
H.~{Ziaei Nafchi} and M.~Cheriet, ``A note on efficiency of downsampling and
  color transformation in image quality assessment,'' \emph{CoRR}, vol.
  abs/1606.06152, 2016. [Online]. Available:
  \url{http://arxiv.org/abs/1606.06152}
\BIBentrySTDinterwordspacing

\bibitem{Wangsite}
\url{https://ece.uwaterloo.ca/~z70wang/research/ssim/}.

\end{thebibliography}

\end{document}